\documentclass[10pt]{article}
\usepackage[utf8]{inputenc}
\usepackage{authblk}
\usepackage{amsfonts}
\usepackage{graphicx}
\usepackage{bm}
\usepackage{tikz}
\usepackage{enumitem}
\usepackage[authoryear]{natbib}
\usepackage{multirow}
\usepackage{amsmath}
\usepackage{amsthm}
\usepackage{amssymb}
\usepackage{algorithm}
\usepackage{algpseudocode}
\usepackage{hyperref}
\hypersetup{
    colorlinks=true,
    citecolor=blue,
    urlcolor=blue,
    linkcolor=blue
}
\usepackage{rotating}
\usepackage{makecell}
\usepackage{float}
\usepackage{dirtytalk}
\usepackage{lscape}
\usepackage{appendix}
\newtheorem{theorem}{Theorem}[section]

\newtheorem{lemma}[theorem]{Lemma}
\theoremstyle{remark}
\newtheorem{remark}{\textbf{Remark}}[section]

\theoremstyle{definition}

\newtheorem{example}{Example}[section]

\usepackage[a4paper, total={6in, 9in}]{geometry}
\usepackage{subcaption}

\DeclareMathOperator*{\argmin}{arg\,min}

\title{rSDNet: Unified Robust Neural Learning against Label Noise and Adversarial Attacks}

\author[1]{\Large{Suryasis Jana}}
\author[1]{\Large{Abhik Ghosh\footnote{Corresponding author: abhik.ghosh@isical.ac.in}}}
\affil[1]{Indian Statistical Institute, Kolkata, India}

\date{}

\begin{document}
\maketitle

\begin{abstract}
Neural networks are central to modern artificial intelligence, yet their training remains highly sensitive to data contamination. 
Standard neural classifiers are trained by minimizing the categorical cross-entropy loss, 
corresponding to maximum likelihood estimation under a multinomial model. 
While statistically efficient under ideal conditions, this approach is highly vulnerable to contaminated observations 
including  label noises  corrupting supervision in the output space, 
and adversarial perturbations inducing worst-case deviations in the input space. 
%Despite their shared impact on empirical risk and generalization, existing methods usually treat these problems separately 
%through specialized correction techniques or computationally expensive adversarial training.
%	
In this paper, we propose a unified and statistically grounded framework for robust neural classification that addresses 
both forms of contamination within a single learning objective. 
We formulate neural network training as a minimum-divergence estimation problem and introduce rSDNet, 
a robust learning algorithm based on the general class of $S$-divergences. 
The resulting training objective inherits robustness properties from classical statistical estimation, 
automatically down-weighting aberrant observations through model probabilities.
We establish essential population-level properties of rSDNet, including Fisher consistency, 
classification calibration implying Bayes optimality, 
and robustness guarantees under uniform label noise and infinitesimal feature contamination. 
Experiments on three benchmark image classification datasets show that rSDNet improves robustness to label corruption 
and adversarial attacks while maintaining competitive accuracy on clean data, 
Our results highlight minimum-divergence learning as a principled and effective framework 
for robust neural classification under heterogeneous data contamination.
\end{abstract}

\bigskip
\noindent\textbf{Keywords:} Neural network classifier; robust learning; $S$-divergences; density power divergence;
adversarial attacks; MLP; CNN.  
\\

% \tableofcontents

\section{Introduction} 
\label{SEC:Intro}

Deep neural networks form the backbone of modern machine learning and artificial intelligence (AI) systems, 
driving advances in vision, language, speech, healthcare, and autonomous decision-making. 
Their success is mainly due to the expressive power and universality of neural network (NN) architectures 
\citep{hornik1989multilayer}, which facilitate flexible modeling of complex data structures. 
Training is commonly performed  via empirical risk minimization on large-scale datasets, 
%most commonly using the cross-entropy (CE) loss for classification, 
implicitly assuming that sample observations faithfully represent the underlying distribution and are free from contamination.
As neural training is increasingly influencing high-stakes AI applications across industry and society, 
ensuring its robustness has become a foundational requirement for the stability, safety, and trustworthiness of modern AI systems.

In practice, however, training data are often contaminated, resulting in significantly poorer performance of neural learning systems.
Two most prevalent sources of such corruption in neural classification are label noise and adversarial perturbations, 
affecting output and input spaces, respectively. Both distort empirical risk compromising the reliability and generalization.
Although traditionally studied separately, they both may be viewed as structured forms of distributional contamination.
This unified perspective motivates the central question of this work: 
can robustness to heterogeneous data deviations be achieved directly through a suitably defined training objective?
Addressing this question, we develop a unified, statistically grounded minimum-divergence learning framework for neural classification, 
termed rSDNet, which provides inherent stability against both types of contamination within a single objective.

Learning with noisy labels has received substantial attention, 
motivated by annotation errors, weak supervision, crowd-sourced labeling, and large-scale web-based data collection. 
Neural classifiers are commonly trained using the categorical cross-entropy (CCE) loss, 
which corresponds to the maximum likelihood (ML) estimation under a multinomial model \citep{Goodfellow-et-al-2016}. 
Due to inherent non-robustness of ML estimation, NNs trained with CCE loss tend to memorize mislabeled samples yielding poor generalization
\citep{natarajan2013learning,zhang2017understanding}. 
Existing solutions include loss correction methods that explicitly model or estimate noise transition mechanisms \citep{patrini2017making}, 
sample selection and reweighting strategies \citep{han2018co,ren2018learning},
and alternative noise-robust loss functions, such as mean absolute error (MAE)  and related losses \citep{ghosh2017robust}, 
generalized cross-entropy (GCE) of \cite{zhang2018generalized}, 
symmetric cross-entropy (SCE) based on the symmetric Kullback–Leibler divergence \citep{wang2019symmetric}, 
trimmed CCE (TCCE) of \cite{rusiecki2019trimmed}, and fractional classification loss (FCL) of \cite{kurucu2025introducing}. 
A comprehensive survey is available in \cite{song2022learning}. 
However, these approaches are primarily tailored to output-space label corruption 
and generally do not address adversarial perturbations in the input space.
%highlighting a structural separation between robustness to corrupted labels and robustness to perturbed inputs in current neural learning systems.

Adversarial robustness addresses a structurally related vulnerability, namely small, carefully crafted input perturbations 
that induce high-confidence misclassification while remaining nearly imperceptible \citep{szegedy2013intriguing,goodfellow2014explaining}. 
From a learning-theoretic perspective, such perturbations represent worst-case local deviations from the nominal data distribution, 
thereby constituting contamination in feature space. Adversarial training remains the dominant defense strategy, 
explicitly optimizing model parameters against worst-case perturbations within a prescribed norm ball \citep{madry2017towards}. 
However, this approach typically incurs substantial computational overhead and may often reduce clean-data accuracy  
\citep{zhang2019theoretically,rice2020overfitting}. 
Related line of works includes certified robustness guarantees and distributionally robust optimization under bounded perturbations 
\citep{sinha2018certifying,cohen2019certified,zhang2019theoretically}. 
%Widely used white-box attack generation algorithms include the Fast Gradient Sign Method (FGSM) of \cite{goodfellow2014explaining}, 
%Projected Gradient Descent (PGD) of \cite{madry2017towards}, the Carlini–Wagner (CW) attack \citep{carlini2017towards}, 
%and DeepFool \citep{moosavi2016deepfool}. 
In this work, instead of attack-specific defenses, we develop a single principled robust loss 
that provides resilience across diverse adversarial perturbation strategies.

More broadly, robust NN learning has evolved along two primary methodological directions. 
The first modifies network architectures or training procedures to reduce sensitivity to corrupted observations.
For example, regularization mechanisms such as dropout \citep{srivastava2014dropout} have been shown 
to improve robustness against label noises in certain settings \citep{rusiecki2020standard}. 
The second adopts a more fundamental statistical viewpoint, constructing loss functions that down-weight aberrant observations,
and is closely related to  distributionally robust optimization against adversarial attacks. 
While these developments highlight the potential of statistically grounded objectives for mitigating contamination effects, 
most of the formal robust loss theory has been  developed in regression contexts 
\citep{rusiecki2007robust,rusiecki2013robust,ghosh2026provably},
while those for classification are often tailored to specific corruption models \citep{qian2022survey}.
Consequently, a common training objective that simultaneously addresses both output- and input-space contamination 
within neural classification remains unexplored.

In this work, we formulate neural classification as a minimum-divergence estimation problem  
and propose a unified robust learning framework, termed rSDNet, based on the general class of $S$-divergences \citep{ghosh2017generalized}. 
The $S$-divergence family provides a two-parameter formulation containing both the $\beta$-divergence \citep{basu1998robust} 
and the Cressie–Read power divergence \citep{cressie1984multinomial}.
It has been formally shown to yield highly efficient and robust statistical inference under potential data contamination  
\citep{ghosh2015asymptotic,ghosh2017minimum}.
By translating these robustness properties to NN classification, 
rSDNet establishes a principled risk minimization strategy in which stability arises 
intrinsically from the curvature and influence down-weighting characteristics of the loss (divergence).
Through theoretical analysis and controlled empirical experiments under both label corruption and adversarial perturbations, 
we demonstrate that robustness can emerge as an inherent property of the learning objective itself, 
rather than as an auxiliary defense mechanism.

%In particular, we establish several key theoretical properties of rSDNet, 
%including Fisher consistency and the classification calibration property.
%We also investigate the theoretical robustness guarantees of rSDNet under both uniform label noises and infinitesimal contamination.
%%and provide practical guidance for selecting appropriate combinations of the tuning parameters. 
%Extensive empirical studies on image classification tasks using three widely used benchmark datasets further support our theoretical findings. They together demonstrate improved robustness of rSDNet under both types of data contamination 
%We further empirically demonstrated that the rSDNet converges faster than the existing learning algorithms under contaminated data with uniform label noise. In addition, the performance of the rSDNet algorithm is evaluated under various types of adversarial attacks.

The rest of the paper is organized as follows.
The proposed rSDNet framework is developed in Section \ref{SECL:proposed-method} along with necessary notations and assumptions. 
In Section \ref{SEC:th-properties}, we establish key theoretical properties of rSDNet, 
including Fisher consistency and the classification calibration property justifying its Bayes optimality.
Here we also investigate the theoretical robustness guarantees of rSDNet under both uniform label noises and infinitesimal contamination.
Empirical studies on image classification tasks using three widely used benchmark datasets are presented in Section \ref{SEC:numericals},
which further support our theoretical findings. Finally, some concluding remarks are given in Section \ref{conclusion}. 
All technical proofs, a discussion on the convergence rate of the proposed rSDNet algorithm, 
and detailed results from our numerical experiments are deferred to Appendices \ref{proofs}--\ref{APP:tabs}.

\section{The proposed robust learning framework}
\label{SECL:proposed-method}

\subsection{Model setup and notations}\label{SEC:setup}

Consider a $J$-class classification problem with a set of $n$ independent training observations 
$\mathcal{S}_n = \{(\bm{y}_i, \bm{x}_i):i=1,2,\ldots,n\}$, 
where $\bm{x}_i\in\mathcal{X} \subseteq \mathbb{R}^p$ denotes the $i$-th input feature vector and 
$\bm{y}_i = (y_{i1},\ldots,y_{iJ})^\top$ is the one-hot encoded categorical response vector indicating the class label of $\bm{x}_i$, 
for $i=1,\ldots,n$. Thus, $\bm{y}_i \in \mathcal{Y} = \{\bm{e}_1,\bm{e}_2,\ldots,\bm{e}_J\}$, 
where $\bm{e}_j$ denotes the $j$-th canonical basis vector in $\mathbb{R}^J$, 
and $\bm{y}_i = \bm{e}_j$ if and only if $\bm{x}_i$ belongs to class $j$, for each $i, j$.
If we denote the random class label corresponding to $\bm{x}_i$ by $Y_i\in\{1, 2, \ldots, J \}$, 
then we can also write $\bm{y}_i = \bm{e}_{Y_i}$ for each $i\geq 1$.
Let us assume that the sampled observations $(\bm{y}_i, \bm{x}_i)$ are independent and identically distributed (IID) realizations of 
the underlying random vectors $(\bm{Y}, \bm{X})$, 
%where $\bm{X}\sim G_{\bm{X}}$ with support $\mathcal{X}$.Further, 
and the conditional distribution of the one-hot encoded response $\bm{Y}$, given $\bm{X}=\bm{x}$, 
is Multinomial$(1; p_{1}^\ast,\ldots, p_{J}^\ast)$, where $p_{j}^*(\bm{x}) = P(\bm{Y} =  \bm{e}_j | \bm{X}=\bm{x})$ 
denotes the posterior probability of class $j=1, \ldots, J$. 
Throughout we assume that, given any $\bm{x}\in\mathcal{X}$, 
 $\bm{p}^*(\bm{x}) = (p_1^\ast(\bm{x}), \ldots, p_J^*(\bm{x}))^\top\in \Delta_J^\circ$, 
the interior of the probability simplex 
$$
\Delta_J = \left\{ \bm{p}=(p_1, \ldots, p_J)^\top \in[0,1]^J : \sum_{j=1}^J p_j =1 \right\}.
$$

Our objective is to model the relationship between class probabilities and input features to facilitate 
classification of new observations. In neural classification, we employ an NN model with $J$ output nodes and a softmax activation
to model the posterior class probabilities $p_j^*(\bm{x})$ by  
\begin{eqnarray}
	p_j(\bm{x}; \bm{\theta}) = \frac{e^{z_j(\bm{x}; \bm{\theta})}}{\sum_{k=1}^J e^{z_k(\bm{x}; \bm{\theta})}}, ~~~~~j=1,\ldots,J,
	\label{EQ:NN-model}
\end{eqnarray}
where $z_j(\bm{x};\bm{\theta})$ denotes the pre-activation (net input) at the $j$-th output node, 
and $\boldsymbol{\theta}\in\Theta$ is the unknown model parameter consisting of  all network weights and biases.
The parameter space $\Theta$ depends on the assumed NN architecture. 
This  model conditional distribution of $\bm{Y}$ given $\bm{X}$ is also Multinomial 
with class probabilities $\bm{p}(\bm{x};\bm{\theta}) = (p_1(\bm{x}; \bm{\theta}),\ldots,p_J(\bm{x}; \bm{\theta}))^\top$,
so that the model probability mass function (PMF) is given by
\begin{equation}\label{EQ:pmf}
	f_{\bm{\theta}}(\bm{u}|\bm{x}) = \prod_{j=1}^J p_j^{u_j}(\bm{x};\bm{\theta}), 
	~~~~ ~ \bm{u}=(u_1, \ldots, u_J)^\top \in \mathcal{Y}.
\end{equation}
If we denote an estimator of $\bm{\theta}$ obtained from the training sample $\mathcal{S}_n$ by $\widehat{\bm{\theta}}_n$,
the resulting plug-in classification rule (NN classifier) is given by 	
$\delta_n(\bm{x}) = \arg \max\limits_{1\leq j\leq J} p_j(\bm{x};\widehat{\bm{\theta}}_n)$.

In practice, the parameter vector $\bm{\theta}$ for an NN classifier is commonly estimated by minimizing the CCE loss function given by 
\begin{equation}\label{CCE}
	\mathcal{L}_{0} (\bm{\theta}) = -\frac{1}{n} \sum_{i=1}^n \sum_{j=1}^J y_{ij} \ln{p_j(\bm{x}_i;\bm{\theta})},
\end{equation}
which coincides with the ML estimation of the model parameters under the multinomial PMF in (\ref{EQ:pmf}). 
Consequently, the resulting classifier inherits the well-known sensitivity 
of ML methods to any form of data contamination and model misspecification.

%\subsection{Neural Network Training as a Minimum Divergence Problem}
\subsection{Minimum divergence learning framework}\label{SEC:MDE}

The equivalence between CCE-based training and ML estimation naturally places neural classification 
within the broader framework of minimum divergence estimation (MDE). 
Statistical MDE provides a principled approach to parameter estimation, 
where estimators are obtained by minimizing a divergence between the empirical distribution and a parametric model,
with the choice of divergence determining their properties.
Historically, divergence-based estimation can be traced back to Pearson’s chi-squared divergence, 
one of the earliest theoretically grounded approaches to statistical inference. 
%ML estimation itself is a special case of MDE corresponding to the Kullback–Leibler divergence (KLD). 
In recent decades, divergence-based methods have received renewed attention due to their ability to produce robust estimators 
with little or no loss in pure data efficiency when an appropriate divergence is selected; see, e.g, \cite{basu2011statistical}.

To formalize this framework in the present setting of NN classification (Section \ref{SEC:setup}), 
let $g(\bm{u}|\bm{x})$, $\bm{u}\in\mathcal{Y}$, denotes the true conditional PMF of $\bm{Y}$ given $\bm{X}=\bm{x}\in\mathcal{X}$, 
corresponding to the Multinomial distribution with true class probabilities $\bm{p}^\ast(\bm{x})$, for each $i=1, \ldots, n$. 
Its empirical counterpart at the observed feature values is directly obtained as 
$\widehat{g}_i(\bm{u}):= \widehat{g}(\bm{u}|\bm{x}_i)=\mathbb{I}(\bm{u}=\bm{y}_i)$ for $\bm{u}\in\mathcal{Y}$ and $i=1, \ldots, n$,
where $\mathbb{I}(\cdot)$ denotes the indicator function. 
%In other words, $\widehat{g}_i$ is a degenerate PMF placing unit mass at the observed label $\bm{y}_i$.
Without assuming any model distribution for $\bm{X}$, here we are basically modelling each conditional PMF $g_i = g(\cdot|\bm{x}_i)$ 
by the parametric PMF $f_{i, \bm{\theta}} = f_{\bm{\theta}}(\cdot|\bm{x}_i)$, as given in (\ref{EQ:pmf}), 
over observed feature values $\bm{x}_i$, $i=1, \ldots, n$.
Thus, a general minimum divergence estimator  $\widehat{\bm{\theta}}_n$ of $\bm{\theta}$, 
with respect to a statistical divergence measure $d(\cdot, \cdot)$, is defined as
$
\widehat{\bm{\theta}}_n =  \argmin_{\bm{\theta}} \frac{1}{n}\sum_{i=1}^n d\bigl(\widehat{g}_i, f_{i,\bm{\theta}}\bigr).
$
%where $d(\cdot,\cdot)$ is a generic statistical divergence between two PMFs defined on the same support $\mathcal{Y}$.

In particular, the ML estimation corresponds to the MDE based on the Kullback--Leibler divergence (KLD) $d_{\mathrm{KL}}(\cdot, \cdot)$. 
%given by defined by$d_{\mathrm{KL}}(g,h) = \sum\limits_{\bm{u}\in\mathcal{Y}}g(\bm{u}) \log \frac{g(\bm{u})}{h(\bm{u})}$ for 
%two PMFs $g$ and $h$ with common support $\mathcal{Y}$.
A straightforward calculation yields 
\[
d_{\mathrm{KL}}(\widehat{g}_i, f_{i,\bm{\theta}})
= \sum\limits_{\bm{u}\in\mathcal{Y}}\widehat{g}_i(\bm{u}) \log \frac{\widehat{g}_i(\bm{u})}{f_{i, \bm{\theta}}(\bm{u})}
= - \sum_{j=1}^J y_{ij}\log p_j(\bm{x}_i;\bm{\theta}),
%+ \sum_{\bm{u}\in\mathcal{Y}}\widehat{g}_i(\bm{u}) \log \widehat{g}_i(\bm{u}),
\]
since $\widehat{g}_i(\bm{e}_j) = y_{ij}$ for all $i, j\geq 1$, and hence 
$\sum_{\bm{u}}\widehat{g}_i(\bm{u}) \log \widehat{g}_i(\bm{u})=\sum_j y_{ij}\log y_{ij} = 0$ under one-hot encoding 
(with the convention $0\cdot \log 0 =0$). 
Consequently, the CCE loss in (\ref{CCE}) can be expressed  as the average KLD measure $
%\mathcal{L}_0(\bm{\theta})= C + 
\frac{1}{n}\sum_{i=1}^n d_{\mathrm{KL}}(\widehat{g}_i, f_{i,\bm{\theta}})$,
%where $C$ is independent of $\bm{\theta}$ (possibly zero). 
Hence, minimizing the CCE loss is exactly equivalent to minimizing the average KLD between the empirical and model conditional PMFs.

This characterization motivates a general minimum divergence learning framework for neural classifiers, 
where the KLD may be replaced by a suitably chosen alternative divergence to achieve desirable statistical properties. 
In particular, divergences possessing suitable robustness properties can substantially mitigate 
the sensitivity of ML-based training to contamination and model misspecification.
Here we adopt the particular class of $S$-divergences \citep{ghosh2017generalized} to construct our proposed rSDNet,
yielding enhanced robustness under both input- and output-space contamination 
while retaining high statistical efficiency under clean data.

\subsection{$S$-Divergence family and the rSDNet objective}

As introduced by \cite{ghosh2017generalized}, the $S$-divergence (SD) between two PMFs, $g$ and $f$ having common support $\mathcal{Y}$ 
is defined, depending on two tuning parameters $\beta\geq 0$ and $\lambda\in\mathbb{R}$, as
%and is given by
\begin{equation*}\label{EQ:SD-def}
S_{\beta,\lambda}(g,f) = \frac{1}{A} \sum_{\bm{u}\in \mathcal{Y}} f^{1+\beta}(\bm{u}) 
- \frac{1+\beta}{AB} \sum_{\bm{u}\in \mathcal{Y}} f^B(\bm{u}) g^A(\bm{u})  + \frac{1}{B} \sum_{\bm{u}\in \mathcal{Y}} g^{1+\beta}(\bm{u}), 
%~~~~\beta\in [0,1], ~\lambda\in\mathbb{R},
\end{equation*}
where $A = 1+\lambda(1-\beta)\neq 0$, and $B = \beta-\lambda(1-\beta)\neq 0$. 
When either $A = 0$ or $B = 0$, the corresponding SD measures are defined through respective continuous limits.
% and have the form
%\begin{equation*}
%	S_{\beta,\lambda:A=0}(g,f) = \lim_{A\to 0} S_{\beta,\lambda}(g,f) = \sum_{y\in\mathcal{Y}} f^{1+\beta}(y) \ln{\left(\frac{f(y)}{g(y)}\right)}  - \frac{1}{1+\beta} \sum_{y\in\mathcal{Y}} \left(f^{1+\beta}(y) - g^{1+\beta}(y)\right),
%\end{equation*}
%%and
%\begin{equation*}
%	S_{\beta,\lambda:B=0}(g,f) = \lim_{B\to 0} S_{\beta,\lambda}(g,f) = \sum_{y\in\mathcal{Y}} g^{1+\beta}(y) \ln{\left(\frac{g(y)}{f(y)}\right)}   - \frac{1}{1+\beta} \sum_{y\in\mathcal{Y}} \left(g^{1+\beta}(y) - f^{1+\beta}(y)\right).
%\end{equation*}
%
The tuning parameters $(\beta,\lambda)$ control the robustness–efficiency trade-off of the resulting MDE and associated inference. 
Particularly, $\beta$ adjusts the influence of outlying observations, so that larger $\beta$ increases robustness but reduces efficiency; 
values of $\beta>1$ are typically avoided due to severe efficiency loss.
On the other hand, $\lambda$ interpolates between divergence families allowing further (higher-order) control over robustness  
while preserving the same first-order asymptotic efficiency at a fixed $\beta$. 
However, \cite{ghosh2017generalized} demonstrated that the minimum SD estimators (MSDEs) exhibit good robustness only when $A \geq 0$. 
Recently, \cite{roy2026asymptotic} established the asymptotic breakdown point of the MSDEs and associated functionals,
confirming high-robustness when $A\geq 0$ and $B\geq 0$. 
Both studies further suggest that the best robustness–efficiency trade-offs are obtained for appropriate SD measures with $A>0$ and $B>0$.
Thus, excluding the boundary cases $A=0$ or $B=0$, here we restrict ourselves to the SDs with admissible tuning parameters 
satisfying $A>0$ and $B>0$, namely  
$$
\mathcal{T} = \left\{ (\beta, \lambda) : -\frac{1}{1-\beta}<\lambda<\frac{\beta}{1-\beta},  0\leq \beta< 1 \right\}
\cup\bigg\{(1, \lambda) : \lambda\in\mathbb{R}\bigg\}.
$$

%\begin{eqnarray}
%0\leq \beta \leq 1, ~~~~\mbox{and }~~~\lambda > - \frac{1}{1-\beta}.
%\label{EQ:tp_values}
%\end{eqnarray}
%
%\begin{equation}\label{dpd-def}
%	d_\beta(g,f) =
%	\begin{cases}
	%		\sum_{y\in\mathcal{Y}} \left\{f^{1+\beta}(y) - \left(1+\frac{1}{\beta}\right)f^\beta(y) g(y) + \frac{1}{\beta}g^{1+\beta}(y)\right\}  , & \mbox{if}~~~ \beta>0,\\
	%		\sum_{y\in\mathcal{Y}} g(y) \ln\left(\frac{g(y)}{f(y)}\right)  , & \mbox{if}~~~ \beta = 0,
	%	\end{cases}
%\end{equation}

Note that the second part of $\mathcal{T}$ corresponds to a single divergence, 
since the SD reduces to the squared $L_2$ distance at $\beta=1$, irrespective of $\lambda$.
Despite this restriction, $\mathcal{T}$ retains most important subclasses of the SD family.
The choice $\lambda=0$ produces $\beta$-divergences or density power divergences (DPDs) of  \cite{basu1998robust},
while $\beta=0$ gives the power divergence (PD) family \citep{cressie1984multinomial} with $-1<\lambda<0$, 
including the Hellinger disparity at $\lambda=-0.5$. However, the set $\mathcal{T}$ exclude the non-robust KLD at $(\beta, \lambda)=(0,0)$ 
and the reverse KLD (rKLD) at $(\beta, \lambda)=(0,-1)$, 
the latter known to cause computational difficulties in discrete models with inliers \citep{basu2011statistical}.
More broadly, the SD family can be viewed as a reparameterization of $(\alpha, \beta)$-divergences \citep{cichocki2011generalized} 
and also as a special case of extended Bregman divergences of \cite{basak2022extended}.

Under our setup of neural classification given in Section \ref{SEC:setup}, 
and following the general discussions in Section \ref{SEC:MDE}, 
we define the MSDE of the NN model parameter $\boldsymbol{\theta}$  as
\begin{equation}
\widehat{\bm{\theta}}_n^{(\beta, \lambda)} = \arg\min_{\bm{\theta}\in\Theta} \mathcal{L}_{\beta,\lambda}^{(n)}(\bm{\theta}) ,
~~~~\mbox{ with } ~~
\mathcal{L}_{\beta,\lambda}^{(n)}(\bm{\theta}) = \frac{1}{n} \sum_{i=1}^n S_{\beta,\lambda}(\widehat{g}_i, f_{i,\bm{\theta}}).
	\label{EQ:SD-MDE-estimator}
\end{equation}
We refer to the resulting neural classifier trained via an MSDE $\widehat{\bm{\theta}}_n^{(\beta, \lambda)}$ 
as rSDNet$(\beta,\lambda)$ for any given choice of $(\beta, \lambda)\in\mathcal{T}$, 
and the associated objective function $\mathcal{L}_{\beta,\lambda}^{(n)}(\bm{\theta})$ as \textit{rSDNet loss} (\textit{empirical SD-risk})
or \textit{rSDNet training objective}.
We next study this loss function further to develop a scalable practical implementation of rSDNet.

\subsection{The final rSDNet learning algorithm}\label{SEC:rSDNet}

For practical implementation of rSDNet, the associated NN training objective $\mathcal{L}_{\beta,\lambda}^{(n)}(\bm{\theta})$ 
can be expressed in a much simplified form. Substituting the explicit forms of $\widehat{g}_i$ and $f_{i,\bm{\theta}}$ %from (\ref{EQ:pmf}) 
into the SD expression (\ref{EQ:SD-def}), %and adjusting for $\bm{\theta}$-independent constants suitably, 
we obtain the final rSDNet training objective (loss) as given by
\begin{equation}\label{EQ:SD-CCE}
    \mathcal{L}_{\beta,\lambda}^{(n)}(\bm{\theta}) = \frac{1}{n} \sum_{i=1}^n \ell_{\beta,\lambda}(\bm{y}_i, \bm{p}(\bm{x}_i;\bm{\theta})),
    ~~~~~(\beta, \lambda)\in\mathcal{T},
\end{equation}
%where
\begin{equation}\label{EQ:SD-loss-pis}
\mbox{where }	~~~\ell_{\beta,\lambda}(\bm{u}, \bm{p})
= \displaystyle \frac{1}{A}\sum_{j=1}^J \left[ p_j^{1+\beta} - \frac{1+\beta}{B} u_j p_j^B + \frac{A}{B} \right],  
~~~\mbox{ for }~(\beta, \lambda)\in\mathcal{T}, ~\bm{u}\in\mathcal{Y}, ~\bm{p}\in\Delta_J.~~~
%	= \left\{\begin{array}{ll}
%    \displaystyle \sum_{j=1}^J \left[ p_j^{1+\beta}(\bm{x}_i;\bm{\theta})
%	- \frac{1+\beta}{B} y_{ij} p_j^B(\bm{x}_i;\bm{\theta}) \right]
%	  + \frac{A}{B},  & \mbox{if }~\lambda\neq \frac{\beta}{1-\beta}, ~\beta\neq 1,\\
%	% -\displaystyle (1+\beta)\left[\sum_{j=1}^J y_{ij}\ln p_j(\bm{x}_i;\bm{\theta}) + \frac{1}{1+\beta}\left(1 - \sum_{j=1}^J p_j^{1+\beta}(\bm{x}_i;\bm{\theta})\right) \right],   & \mbox{if }~\lambda = \frac{\beta}{1-\beta}, ~\beta\neq 1,\\
%    \displaystyle\sum_{j=1}^J \left[p_j^{1+\beta} (\bm{x}_i;\bm{\theta}) - (1+\beta) y_{ij}\ln p_j(\bm{x}_i;\bm{\theta}) \right] - 1,   & \mbox{if }~\lambda = \frac{\beta}{1-\beta}, ~\beta\neq 1,\\
%	\displaystyle \sum_{j=1}^J p_j^2(\bm{x}_i;\bm{\theta}) - 2\prod_{j=1}^J p_j^{y_{ij}}(\bm{x}_i;\bm{\theta}) + 1,   & \mbox{if }~\beta= 1.
%	\end{array}\right.
\end{equation}
Although we restrict ourselves to tuning parameter values in $\mathcal{T}$, 
the rSDNet loss $\mathcal{L}_{\beta,\lambda}(\bm{\theta})$ can also be extended to the boundary cases of $A=0$ or $B=0$ 
by the respective continuous limits of the form given in \eqref{EQ:SD-CCE}--(\ref{EQ:SD-loss-pis}). 
We avoided these boundary cases as they are expected to have practical issues with either outliers or inliers \citep{ghosh2017generalized}.
%The additive constant $\frac{A}{B}$ is kept there in \eqref{EQ:SD-loss-pis} to ensure that its limit, as $B\rightarrow 0$, exists
%and equals to the expression presented under the case $B=0$, i.e., $\lambda = \beta/(1-\beta)$. 
%The loss function becomes independent of $\lambda$ at $\beta=1$ as expected.
%The MSDE $\widehat{\bm{\theta}}_n^{(\beta, \lambda)}$ in \eqref{EQ:SD-MDE-estimator} can then be obtained 
%by empirically minimizing the simplified loss function $\mathcal{L}_{\beta,\lambda}(\bm{\theta})$ given in \eqref{EQ:SD-CCE} 
%with respect to the NN model parameters $\bm{\theta}\in\Theta$. 

Since the rSDNet objective $\mathcal{L}_{\beta,\lambda}(\bm{\theta})$ in \eqref{EQ:SD-CCE} can be non-convex in $\bm{\theta}$
by the choice of the network architecture in $\bm{p}(\bm{x}; \bm\theta)$, following standard NN learning procedures, 
we propose to solve this optimization problem efficiently using the Adam algorithm \citep{kingma2014adam}.
It is a first-order stochastic gradient method, which starts with initializing two moment vectors $\bm{m}_0$ and $\bm{v}_0$ 
to the null vector, and update the minimizer of $\mathcal{L}_{\beta,\lambda}(\bm{\theta})$  at the $t$-th step of iteration by
\begin{equation} \label{adam-upd}
    \bm{\theta}_t \leftarrow \bm{\theta}_{t-1} - \alpha\ \bm{\widehat{m}}_t/\left(\sqrt{\bm{\widehat{v}}_t} + \epsilon \right),
    ~~~~ t=1, 2, \ldots, 
\end{equation} 
where $\bm{\widehat{m}}_t$ and $\bm{\widehat{v}}_t$ are, respectively, the updated bias-corrected estimates 
of first and second raw moments, given by
\begin{equation*}
    \bm{\widehat{m}}_t \leftarrow \frac{\beta_1 \bm{m}_{t-1} + (1-\beta_1)\bm{g}_t}{1-\beta_1^t},~~ 
    \bm{\widehat{v}}_t \leftarrow \frac{\beta_2 \bm{v}_{t-1} + (1-\beta_2)\bm{g}_t^2}{1-\beta_2^t},
\end{equation*}
with $\bm{g}_t = \nabla_{\bm{\theta}} \mathcal{L}_{\beta,\lambda}(\bm{\theta}_{t-1})$
denoting the gradient of the loss function with respect to $\bm{\theta}$, and  
$\bm{g}_t^2$ being the vector of squared elements of $\bm{g}_t$ for each $t\in\mathbb{N}$.
%\begin{equation*}
%    \bm{m}_t \leftarrow\beta_1 \bm{m}_{t-1} + (1-\beta_1)\bm{g}_t,~~
%    \bm{v}_t \leftarrow\beta_1 \bm{v}_{t-1} + (1-\beta_2)\bm{g}_t^2,~~
%    \bm{g}_t = \nabla_{\bm{\theta}} L_{\beta,\lambda}^{SD}(\bm{\theta}_{t-1}).
%\end{equation*}
%Here $\nabla_{\bm{\theta}}$ denotes the gradient with respect to $\bm{\theta}$ 
Given the form of the rSDNet loss function in (\ref{EQ:SD-CCE})--(\ref{EQ:SD-loss-pis}), 
we can  compute its gradient  as given by 
%and has the simplified form (under one-hot encoding) 
\begin{equation}
\nabla_{\bm{\theta}} \mathcal{L}_{\beta,\lambda}(\bm{\theta})
= \displaystyle\frac{1+\beta}{nA} \sum_{i=1}^n \sum_{j=1}^J
\left[p_j^{\beta}(\bm{x}_i;\bm{\theta}) - y_{ij} p_j^{B-1}(\bm{x}_i;\bm{\theta})  \right] \nabla_{\bm{\theta}}p_j(\bm{x}_i;\bm{\theta}),
    ~~~~~(\beta, \lambda)\in\mathcal{T}.
%	= \left\{\begin{array}{ll}
%		\displaystyle\frac{1+\beta}{n}
%		\sum_{i=1}^n \sum_{j=1}^J
%		\left[
%		 p_j^{\beta}(\bm{x}_i;\bm{\theta}) 	-
%%	\left(\sum_{j=1}^J y_{ij} p_j(\bm{x}_i;\bm{\theta})\right)^{B-1} 
%	y_{ij} p_j^{B-1}(\bm{x}_i;\bm{\theta})  \right] \nabla_{\bm{\theta}}p_j(\bm{x}_i;\bm{\theta}),  & \mbox{if }~\lambda\neq \frac{\beta}{1-\beta}, ~\beta\neq 1, \\
%	\displaystyle\frac{1+\beta}{n} \sum_{i=1}^n \sum_{j=1}^J \left[ p_j^\beta(\bm{x}_i;\bm{\theta}) - y_{ij}/p_j(\bm{x}_i;\bm{\theta})\right] \nabla_{\bm{\theta}} p_j(\bm{x}_i;\bm{\theta}), & \mbox{if }~\lambda = \frac{\beta}{1-\beta}, ~\beta\neq 1,\\
%    \displaystyle\frac{2}{n}\sum_{i=1}^n\sum_{j=1}^J \left[p_j(\bm{x}_i;\bm{\theta}) - y_{ij}\right] \nabla_{\bm{\theta}}p_j(\bm{x}_i;\bm{\theta}), & \mbox{if }~\beta= 1.
%	\end{array}\right.
	\label{EQ:SD-CCE-grad}
\end{equation}
%\begin{eqnarray}
%\mbox{ with }~~ \nabla_{\bm{\theta}} p_j(\bm{x}; \bm{\theta}) = p_j(\bm{x}; \bm{\theta}) \left[\nabla_{\bm{\theta}}z_j(\bm{x}_t; \bm{\theta}) 
%- \sum_{k=1}^J p_j(\bm{x}; \bm{\theta})\nabla_{\bm{\theta}} z_k(\bm{x}_t; \bm{\theta})\right],	
%~~~j=1, \ldots, J.
%\label{EQ:grad-p}
%\end{eqnarray}
For NN architectures involving non-smooth activation functions (e.g., ReLU), 
its output $ p_j(\bm{x}_i;\bm{\theta})$ might not be differentiable everywhere with respect to $\bm{\theta}$.
In such cases, we need to replace $\nabla_{\bm{\theta}} p_j(\cdot;\bm{\theta})$ in gradient computation 
by a measurable selection from its subdifferential $\partial_{\bm{\theta}}p_j(\cdot,\bm{\theta})$,
which is assumed to exist for most commonly used NN architectures.

For efficient implementation of the Adam algorithm, we have used the \texttt{TensorFlow} library \citep{abadi2015tensorflow} of Python,; 
it computes the required gradient $\bm{g}_t$ via automatic differentiation \citep{bolte2021conservative}, 
covering the cases of non-smooth activation as well. 
As initial weights $\bm{\theta}_0$ in the training process, we have used either of the standard  built-in initializers `Glorot' and `He', 
which correspond to the normal and uniform base distributions, respectively. 
Throughout our empirical experimentation, we have used the default Adam hyperparameter values suggested by \cite{kingma2014adam},
which are $\alpha=10^{-3}$, $\beta_1=0.9$, $\beta_2=0.999$, and $\epsilon=10^{-8}$.
Python codes for the complete rSDNet training are available through the GitHub repository 
\texttt{Robust-NN-learning}\footnote{\url{https://github.com/Suryasis124/Robust-NN-learning.git}}.

Following the theory of Adam \citep{kingma2014adam}, one should ideally continue the iterative updation step \eqref{adam-upd} 
until the sequence $\{\bm{\theta}_t\}$ converges. However, in most practical problems, 
achieving such full convergence is often computationally expensive.
So, Adam updates are typically executed for a predefined number of epochs,
which is generally chosen by monitoring the stability of the test accuracy for the trained classifier. 
In all illustration presented in this paper, we employed the Adam algorithm for 250 epochs, 
which was observed to be sufficient for stable training of rSDNet; 
see Appendix \ref{APP:conv} for an empirical study justifying it for both pure and contaminated data. 

%\newpage 	
\section{Theoretical guarantees}\label{SEC:th-properties}

\subsection{Statistical consistency of rSDNet functionals}
\label{SEC:FisherC}

We start by establishing that the proposed rSDNet classifier indeed defines a valid statistical classification rule at the population level,
which is essential for justifying its use in learning from random training  samples under ideal conditions.
In this respect, we need to define the population-level functionals associated with rSDNet,
which characterizes the target parameters that empirical learning procedures aim to estimate.

We may note that the MSDE $\widehat{\bm{\theta}}_n^{(\beta, \lambda)}$ of the NN model parameter $\boldsymbol{\theta}$ under 
rSDNet$(\beta,\lambda)$ may be re-expressed from \eqref{EQ:SD-MDE-estimator} as a minimizer of 
$E_{G_{\bm{X},n}}[S_{\beta, \lambda}(\widehat{g}(\cdot|\bm{X}), f_{\bm{\theta}}(\cdot|\bm{X}))]$
with respect to $\bm{\theta}\in\Theta$, 
where $G_{\bm{X},n}$ denotes the empirical distribution function of $\bm{X}$ placing mass $1/n$ at each observed $\bm{x}_i$, 
and $\widehat{g}$ is the empirical estimate of $g$ based on the training sample $\mathcal{S}_n$. 
Accordingly, we define the population-level \textit{minimum SD functional} (MSDF) of $\bm{\theta}$ 
for tuning parameters $(\beta, \lambda)\in\mathcal{T}$ at  the true distributions $(G_{\bm{X}}, g)$ as 
\begin{equation}\label{EQ:rsdnet-func}
\bm{T}_{\beta, \lambda}(G_{\bm{X}}, g) = \argmin\limits_{\bm{\theta}\in\Theta} \mathcal{R}_{\beta,\lambda}(\bm{\theta}|G_{\bm{X}}, g)
~~~\mbox{ with }~~
\mathcal{R}_{\beta,\lambda}(\bm{\theta}|G_{\bm{X}}, g) 
= E_{G_{\bm{X}}}\left[S_{\beta, \lambda}(g(\cdot|\bm{X}), f_{\bm{\theta}}(\cdot|\bm{X}))\right],
\end{equation}
%whenever such a minimum exists. 
where $G_{\bm{X}}$ denotes the true distribution function of $\bm{X}$ and 
$\mathcal{R}_{\beta,\lambda}(\bm{\theta}) = \mathcal{R}_{\beta,\lambda}(\bm{\theta}|G_{\bm{X}}, g)$ is the \textit{population SD-risk}. 
At the empirical level, this corresponds to 
$\mathcal{R}_{\beta,\lambda}(\bm{\theta}|G_{\bm{X},n}, \widehat{g}) = \mathcal{L}_{\beta,\lambda}^{(n)}(\bm{\theta})$, 
so that $\bm{T}_{\beta, \lambda}(G_{\bm{X},n}, \widehat{g})= \widehat{\bm{\theta}}_n^{(\beta, \lambda)}$.
We refer to the resulting neural classifier obtained by setting $\bm{\theta} = \bm{T}_{\beta, \lambda}(G_{\bm{X}}, g)$ 
as the \textit{rSDNet functional}. The following theorem then presents its Fisher consistency; 
the proof is straightforward from the fact that SD is a genuine statistical divergence \citep{ghosh2017generalized}.

\begin{theorem}[Fisher Consistency of rSDNet]  \label{THM:Fisher_consistency}
For any $(\beta, \lambda)\in\mathcal{T}$, the posterior class probabilities of the rSDNet functional,
obtained using the MSDF in \eqref{EQ:rsdnet-func}, satisfy 
\[
\bm{p}(\bm{x}; \bm{T}_{\beta, \lambda}(G_{\bm{X}}, g))  = \bm{p}(\bm{x}; \bm{\theta}_0) = \bm{p}^*(\bm{x})
\quad \text{a.s.},
\]
for any marginal distribution $G_{\bm{X}}$, provided that the conditional model is correctly specified 
with $g \equiv f_{\bm{\theta}_0}$ for some $\bm{\theta}_0 \in \Theta$.
\end{theorem}

The above theorem shows Fisher consistency and uniqueness of the rSDNet functional at the level of class probabilities.
% determined by the outputs of the assumed NN model. 
A similar result for the MSDF $\bm{T}_{\beta, \lambda}(G_{\bm{X}}, g)$ requires 
the following standard identifiability condition for the assumed NN architecture.

\begin{enumerate}[label=(A0), ref=(A0)]
\item \label{itm:AO} 
The NN classifier output function $\bm{x} \mapsto \bm{p}(\bm{x};\bm{\theta})$ is measurable for all $\bm{\theta}\in\Theta$, 
and the parameterization $\bm{\theta} \mapsto \bm{p}(\bm{x};\bm{\theta})$ is identifiable in $\bm{\theta}$ up to known network symmetries, 
i.e.,
	\begin{equation*}
		\bm{p}(\bm{x};\bm{\theta}_1) = \bm{p}(\bm{x};\bm{\theta}_2)~~ a.s.~~~~~
		\Rightarrow ~~~\bm{\theta}_1 = \mathbb{g} \cdot \bm{\theta}_2,
	\end{equation*}
for some $\mathbb{g}$ in a known symmetry group $\mathcal{G}$ acting on the (non-empty) parameter space $\Theta$.
\end{enumerate}

Under Assumption \ref{itm:AO}, the MSDF at any $(\beta, \lambda)\in\mathcal{T}$ satisfies  
$\bm{T}_{\beta, \lambda}(G_{\bm{X}}, g) = \mathbb{g}\cdot \bm{\theta}_0$ for some $\mathbb{g}\in\mathcal{G}$ as  in \ref{itm:AO}. 
Thus, $\bm{T}_{\beta, \lambda}(G_{\bm{X}}, g)$ is unique only up to known NN symmetries,
%since the identifiability of the model parameters with respect to the network outputs depends on these symmetries. This is 
which is consistent with standard NN learning theory \citep[see, e.g.,][]{Goodfellow-et-al-2016,ghosh2026provably}.

%\subsection{Classification calibration}

Beyond Fisher consistency, the rSDNet functional is also \textit{classification-calibrated}, 
meaning that minimization of the population SD-risk $\mathcal{R}_{\beta,\lambda}(\bm{\theta})$ 
yields Bayes-optimal classification decisions at the population level 
\citep{zhang2004statistical,bartlett2006convexity,tewari2007consistency}. 
To verify this, let us further simplify the population SD-risk, using the form of SD given in (\ref{EQ:SD-def}), as  
$
\mathcal{R}_{\beta,\lambda}(\bm{\theta}) = E_{G_{\bm{X}}}\left[r_{\beta, \lambda}(p^*(\bm{X}), p(\bm{X}; \bm{\theta}))\right],
$
where we define the conditional SD-risk 
\begin{equation}\label{EQ:conditional-risk}
	r_{\beta,\lambda}(\bm{p}^*, \bm{p})
	=
	\frac{1}{A}\sum_{j=1}^J
	\left[
	p_j^{1+\beta}
	-\frac{1+\beta}{B}p_j^{B}{p_j^*}^{A}
	+\frac{A}{B}{p_j^*}^{1+\beta}
	\right],
	\qquad \mbox{for }~ (\beta,\lambda)\in\mathcal{T}, ~\bm{p}, \bm{p}^*\in\Delta_J.
\end{equation}
Note that $r_{\beta,\lambda}(\bm{u}, \bm{p}) = \ell_{\beta,\lambda}(\bm{u},\bm{p})$ for any $\bm{u}\in\mathcal{Y}$ and $\bm{p}\in\Delta_J$,
although they differ for any $\bm{p}^*\in\Delta_J^\circ$ unless $(\beta,\lambda)=(0,-1)$.
Further, since \eqref{EQ:conditional-risk} corresponds to the SD between $\bm{p}^*$ and $\bm{p}$, 
it is always non-negative and equals zero if and only if $\bm{p}=\bm{p}^*$.
Consequently, the population SD-risk $\mathcal{R}_{\beta,\lambda}(\bm{\theta})$ has a unique minimizer 
at $\bm{p} = \bm{p}^*$ for any $(\beta,\lambda)\in\mathcal{T}$.
In other words, the SD-loss underlying rSDNet is classification-calibrated in the sense of \cite{bartlett2006convexity}.
Hence, the class predicted by the rSDNet functional coincides with that of the asymptotically optimal Bayes classifier, 
$\delta_B(\bm{x}) = \arg\max_{1\le j\le J} p_j^*(\bm{x})$, which minimizes the conditional misclassification risk.
These properties of rSDNet are summarized in the following theorem.

\begin{theorem}[Classification calibration of rSDNet]
    \label{THM:cl-calibrated}
The rSDNet functional at any $(\beta, \lambda) \in \mathcal{T}$ is classification-calibrated, and hence 
the induced classifier achieves the Bayes-optimal class prediction at the population level.
\end{theorem}

%Theorem \ref{THM:cl-calibrated} establishes that the surrogate loss used by rSDNet is statistically consistent for classification.
%%minimizing the population SD-risk yields a classifier that coincides with the Bayes rule. 
%Thus, with a sufficiently expressive neural network and consistent empirical risk minimization, 
%rSDNet achieves asymptotically optimal classification performance.
%Combined with the Fisher consistency, this result provides theoretical justification 
%for using the SD-loss as a surrogate objective for robust neural classification.
%%In particular, the result shows that the robustness introduced by the SD-loss does not compromise the fundamental requirement of 
%%Bayes optimality, a key property studied in the theory of surrogate loss functions \citep{bartlett2006convexity, tewari2007consistency}.

\subsection{Tolerance against uniform label noise} \label{RAULN}

We now establish the robustness of rSDNet against uniform label noise, a widely used output noise model in classification.
Each training label is assumed to be independently corrupted with probability $\eta\in[0,1]$; 
when corruption occurs, the label is replaced uniformly at random by one of the remaining class labels. 
Formally, we represent contaminated training  data as $\mathcal{S}_\eta = \{(\widetilde{\bm{y}}_i, \bm{x}_i):i=1,2,\ldots,n\}$, 
where the observed label $\widetilde{\bm{y}}_i=\bm{y}_i$ with probability $1-\eta$ 
and equals any incorrect class label with probability $\eta/(J-1)$ when the input feature value is $\bm{x}_i$,  $i\geq 1$. 
If we denote the random variable corresponding to $\widetilde{\bm{y}}_i$ by $\widetilde{\bm{Y}}$,
then the true conditional distribution of $\widetilde{\bm{Y}}$, given $\bm{X}=\bm{x}$, is again Multinomial but with PMF 
$g_\eta(\cdot|\bm{x})$ involving class probabilities $\bm{p}_\eta^*(\bm{x}) = (p_{\eta,1}^{*}(\bm{x}),\ldots,p_{\eta, J}^{*}(\bm{x}))$,  where 
\begin{equation*}
p_{\eta,j}^{*}(\bm{x}) = (1-\eta)p_j^*(\bm{x}) + \frac{\eta}{J-1}(1 - p_j^*(\bm{x})), ~~~~~ j=1,\ldots,J.
\end{equation*}

We study the effect of such noise on the \textit{expected SD-loss}  underlying our rSDNet, given by 
\begin{equation*}
   R_{\beta,\lambda}(\bm{\theta}) = E_{(\bm{Y}, \bm{X})}\left[\mathcal{L}_{\beta,\lambda}^{(n)}(\bm{\theta})\right] 
   = E_{(\bm{Y}, \bm{X})}\left[\ell_{\beta,\lambda}(\bm{Y}, \bm{p}(\bm{X};\bm{\theta}))\right]
   = E_{G_{\bm{X}}}\left[E_{g}\left[\ell_{\beta,\lambda}(\bm{Y}, \bm{p}(\bm{X};\bm{\theta}))|\bm{X}\right]\right].
\end{equation*}
Let $\bm{\theta}_{\beta,\lambda}^*$ denotes its global minimizer at $(\beta, \lambda)\in\mathcal{T}$. 
We compare the minimum achievable expected SD-loss  $R_{\beta,\lambda}(\bm{\theta}_{\beta,\lambda}^*)$ under clean data 
with that obtained under uniform label noise. To this end, we define $\bm{\theta}_{\beta,\lambda}^\eta$ to be the global minimizer 
of the expected SD-risk under uniform label noise, given by 
\begin{equation*}
  R_{\beta,\lambda}^\eta(\bm{\theta}) = E_{(\widetilde{\bm{Y}}, \bm{X})}\left[\mathcal{L}_{\beta,\lambda}^{(n)}(\bm{\theta})\right] 
= E_{G_{\bm{X}}}\left[E_{g_\eta}\left[\ell_{\beta,\lambda}(\bm{Y}, \bm{p}(\bm{X};\bm{\theta}))|\bm{X}\right]\right].
\end{equation*}
Then we assess the robustness of rSDNet through the excess risk, 
$R_{\beta,\lambda}(\bm{\theta}_{\beta,\lambda}^\eta) - R_{\beta,\lambda}(\bm{\theta}_{\beta,\lambda}^*)$. 
Smaller this difference close to zero, greater the tolerance of rSDNet against uniform label noises. 
To derive a bound for it, we first show the uniform boundedness of the total SD-loss in the following lemma;
its proof is given in Appendix \ref{App:sd-loss-bounded}.

\begin{lemma}\label{lem:sd-loss-bounded}
For any $\bm{x}\in\mathcal{X}$, $\bm{\theta}\in\Theta$, and $(\beta, \lambda)\in\mathcal{T}$, 
we have 
\begin{equation}\label{eq:sd-loss-bounds}
\frac{J^{1-\beta}}{A} - \frac{1+\beta}{AB} \max(1, J^{1-B}) + \frac{J^2}{B}
		\;\le\;
		\sum_{j=1}^J \ell_{\beta,\lambda}(\bm{e}_j, \bm{p}(\bm{x}; \bm{\theta}))
		\;\le\;
		\frac{J}{A} - \frac{1+\beta}{AB} \min(1, J^{1-B}) + \frac{J^2}{B}.
	\end{equation}
\end{lemma}

Since the total SD-loss in Lemma \ref{lem:sd-loss-bounded} depends on $(\bm{x},\bm{\theta})$, 
it is not symmetric in the sense of \cite{ghosh2017robust}. Symmetry arises only at $(\beta,\lambda)=(0,-1)$, 
corresponding to the rKLD having $A=0$, which lies outside our admissible parameter range $\mathcal{T}$; 
see \cite{wang2019symmetric} for neural classifiers developed based on the rKLD and its practical limitations.

Although SD-loss is not symmetric and thus not exactly noise-tolerant, 
rSDNet still exhibits bounded excess risk under uniform label noise for all $(\beta,\lambda)\in\mathcal{T}$. 
The exact bound is provided in the following theorem; see Appendix \ref{App:sd-loss-noise-robust} for its proof.

\begin{theorem}\label{thm:sd-loss-noise-robust}
Under uniform label noise with contamination proportion $0\leq\eta < 1-1/J$, 
the excess risk of rSDNet satisfies 
%at any  $(\beta, \lambda)\in\mathcal{T}$ remains bounded with
\begin{eqnarray}
0 \le R_{\beta,\lambda}(\bm{\theta}_{\beta,\lambda}^\eta) -  R_{\beta,\lambda}(\bm{\theta}_{\beta,\lambda}^*) 
\le M_{\beta,\lambda}^{(\eta)},  ~~~~ (\beta, \lambda)\in\mathcal{T},
\label{EQ:label_noise_bound}
\end{eqnarray}
where
\[
M_{\beta,\lambda}^{(\eta)} = \frac{\eta}{(J-1-J\eta)} 	\frac{1}{A} \left( J - J^{1-\beta} + \frac{1+\beta}{B}|1 - J^{1-B}|\right).
\]
\end{theorem}

\begin{figure}[H]
     \centering
     \begin{subfigure}[b]{0.32\textwidth}
         \centering
         \includegraphics[width=\textwidth]{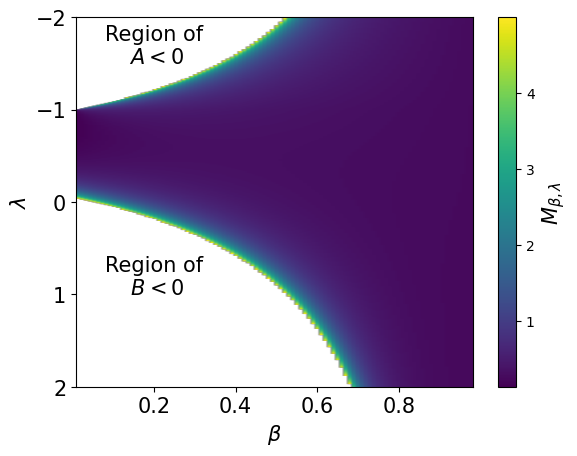}
         \caption{$\eta=0.2$}
         \label{fig:heat_2a}
     \end{subfigure}
%     \hfill
     \begin{subfigure}[b]{0.32\textwidth}
         \centering
         \includegraphics[width=\textwidth]{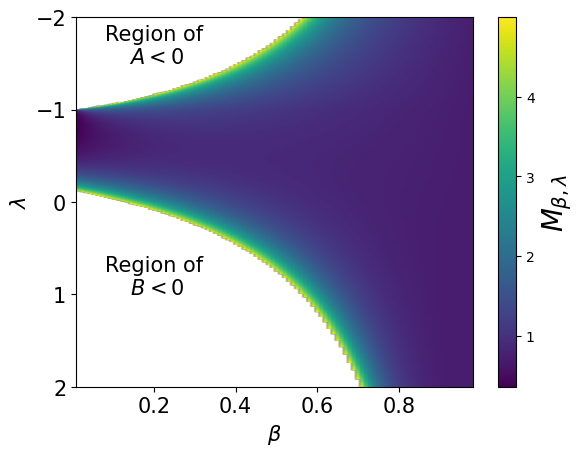}
         \caption{$\eta=0.4$}
         \label{fig:heat_2b}
     \end{subfigure}
     \begin{subfigure}[b]{0.32\textwidth}
	\centering
	\includegraphics[width=\textwidth]{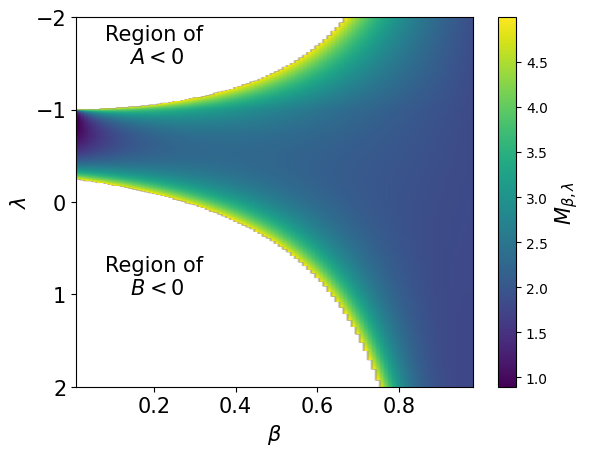}
	\caption{$\eta=0.6$}
	\label{fig:heat_2c}
\end{subfigure}
     \caption{Plots of the bound $M_{\beta,\lambda}^{\eta}$ on the excess risk of rSDNet, 
     	as a function of tuning parameters $(\beta, \lambda)$, 
     	for different contamination proportion $\eta$ and $J=10$.}
     \label{fig:heat_2}
\end{figure}

Note that the bound $M_{\beta,\lambda}^{(\eta)}$ in Theorem \ref{thm:sd-loss-noise-robust} decomposes into two components: 
one depending on the noise level $\eta$ and the other on the tuning parameters $(\beta,\lambda)$. 
The $\eta$-dependent factor coincides with that obtained in robust learning literature \citep[e.g.,][]{ghosh2017robust}, 
implying the same rate of degradation with increasing noise. 
In particular, the bound vanishes as $\eta\downarrow0$ and diverges as $\eta\uparrow (J-1)/J$. 
The second component in $M_{\beta,\lambda}^{(\eta)}$ quantifies the effect of $(\beta,\lambda)$ 
and explains the improved robustness of certain rSDNet members. 
For fixed $(\eta, J)$, the bound decreases as $\beta$ increases, indicating greater tolerance to uniform label noise
(see Figure \ref{fig:heat_2}). The role of $\lambda$ becomes more pronounced at higher contamination levels;
for any fixed $\beta\in[0,1]$, the bound is minimized when $\lambda\in[-1,0]$ 
and increases rapidly as $\lambda$ moves outside this range.

\subsection{Local robustness against contaminated features}
\label{SEC:IF}

We next evaluate the local robustness of rSDNet to infinitesimal contamination in the feature space using influence function (IF) analysis. 
Since sample-based IFs are often difficult to interpret in deep learning settings \citep[see, e.g.,][]{basu2020influence,bae2022if}, 
we adopt the population-level formulation of IF as originally introduced by \cite{hampel1986robust}. 
Specifically, we study the IF of the MSDF $\bm{T}_{\beta,\lambda}(G_{\bm{X}},g)$ underlying our rSDNet, 
under a gross-error contaminated feature distribution $G_{\bm{X},\epsilon} = (1-\epsilon)G_{\bm{X}} + \epsilon\wedge_{\bm{x}_t}$,
where $\epsilon>0$ denotes the contamination proportion at an outlying point $\bm{x}_t\in\mathcal{X}$
and $\wedge_{\bm{x}}$ denotes the degenerate distribution at  $\bm{x}\in\mathcal{X}$.
The IF of $\bm{T}_{\beta,\lambda}$ at the contamination point $\bm{x}_t$ is formally defined as
\begin{equation*}	
	\mathcal{IF}(\bm{x}_t, \bm{T}_{\beta,\lambda}, (G_{\bm{X}}, g)) 
	= \lim\limits_{\epsilon\downarrow 0}\frac{\bm{T}_{\beta, \lambda}(G_{\bm{X}}, g) - \bm{T}_{\beta, \lambda}(G_{\bm{X},\epsilon}, g)}{\epsilon}
	=\frac{\partial}{\partial\epsilon} \bm{T}_{\beta, \lambda}(G_{\bm{X},\epsilon}, g) \bigg|_{\epsilon=0},
\end{equation*}
which  quantifies the effect of infinitesimal contamination at $\bm{x}_t$ on the resulting estimator and hence on the rSDNet framework.

In order to derive the expression of this IF at any $(\beta, \lambda)\in\mathcal{T}$, we define
\begin{eqnarray}
\bm{\psi}_{\beta, \lambda}(\bm{x}; \bm{\theta}) &=&  
\frac{A}{1+\beta} \nabla_{\bm{\theta}} \left[r_{\beta, \lambda}(\bm{p}^*(\bm{x}), \bm{p}(\bm{x}; \bm{\theta}))\right] 
=
\sum_{j=1}^{J} u_j(\bm{x};\bm{\theta}) \nabla_{\bm{\theta}}p_j(\bm{x};\bm{\theta}),
\label{EQ:psi}\\
~~\mbox{with}&&
u_j(\bm{x};\bm{\theta}) = p_j^{\beta}(\bm{x};\bm{\theta}) - {p_j^*}^A(\bm{x})p_j^{B-1}(\bm{x};\bm{\theta}),
\nonumber\\
%\mbox{so that }~ 
%\bm{\xi}_{\beta, \lambda}(\bm{\theta}) &=&  \frac{A}{1+\beta} \nabla_{\bm{\theta}} \mathcal{R}_{\beta, \lambda}(\bm{\theta}) 
%= E_{G_{\bm{X}}} \left[\bm{\psi}_{\beta, \lambda}(\bm{X}; \bm{\theta})\right],
%\label{EQ:xi}\\
\mbox{and }~~~ 
\bm{\Psi}_{\beta,\lambda}(\bm{\theta}) &=& \frac{A}{1+\beta} \nabla_{\bm{\theta}}^2 \mathcal{R}_{\beta, \lambda}(\bm{\theta}) 
= E_{G_{\bm{X}}} \left[\nabla_{\bm{\theta}}\bm{\psi}_{\beta, \lambda}(\bm{X}; \bm{\theta})\right],
\nonumber\\
&= & E_{G_{\bm{X}}}\left[ \sum_{j=1}^{J} u_j'(\bm{X};\bm{\theta}) 
\nabla_{\bm{\theta}}p_j(\bm{X};\bm{\theta})\nabla_{\bm{\theta}}^\top p_j(\bm{X};\bm{\theta}) +
\sum_{j=1}^{J} u_j(\bm{X};\bm{\theta}) \nabla_{\bm{\theta}}^{2}p_j(\bm{X};\bm{\theta}) \right], ~~~~~~~~
\label{EQ:Psi}
\\
\mbox{ with}&&u_j'(\bm{x};\bm{\theta}) = \beta p_j^{\beta-1}(\bm{x};\bm{\theta}) - {p_j^*}^A(\bm{x})(B-1)p_j^{B-2}(\bm{x};\bm{\theta}).
\nonumber
\end{eqnarray}
As noted earlier, for non-differentiable NN architectures, $\nabla_{\bm{\theta}}$ in the above definitions
should be interpreted as a measurable element of the corresponding sub-differential with respect to $\bm{\theta}$. 
Then, the following theorem presents the final IF for the MSDF;
see Appendix \ref{App:sd-loss-noise-robust} for its proof.

\begin{theorem}\label{thm:IF_rSDNet}
For rSDNet with tuning parameter $(\beta,\lambda)\in \mathcal{T}$, the influence function of the underlying MSDF 
$\bm{T}_{\beta,\lambda}$ under feature contamination at $\bm{x}_t\in\mathcal{X}$ is given by 
\begin{align}
\mathcal{IF}(\bm{x}_t,\bm{T}_{\beta,\lambda},(G_{\bm{X}},g))
= - \bm{\Psi}_{\beta,\lambda}^+(\bm{\theta}_g) \bm{\psi}_{\beta, \lambda}(\bm{x}_t, \bm{\theta}_g) 
%-\bm{\xi}_{\beta, \lambda}(\bm{\theta}_g)\right] 
+ \bm{\nu}(\bm{\theta}_g),
\label{EQ:IF-final}
\end{align}
where $\bm{\theta}_g = \bm{T}_{\beta,\lambda}(G_{\bm{X}},g)$, 
$\bm{\nu}(\bm{\theta}_g)\in Ker(\bm{\Psi}_{\beta,\lambda}(\bm{\theta}_g))$, 
the kernel (null-space) of $\bm{\Psi}_{\beta,\lambda}(\bm{\theta}_g)$, 
and $\bm{\Psi}_{\beta,\lambda}^+$ represents Moore–Penrose inverse of $\bm{\Psi}_{\beta,\lambda}$.
\end{theorem}

\begin{remark}
For thin shallow or under-parameterized NN architectures, 
the matrix $\bm{\Psi}_{\beta,\lambda}(\bm{\theta}_g)$ is typically non-singular, 
and so its Moore-Penrose inverse reduces to the ordinary inverse. 
Then, (\ref{EQ:IF-final}) provides the unique IF of the MSDF with $\bm{\nu}(\bm{\theta}_g)=\bm{0}$. 
In more general deep or over-parameterized networks, however, 
$\bm{\Psi}_{\beta,\lambda}(\bm{\theta}_g)$ is often singular, implying non-uniqueness of the IF. 
Nevertheless, the dependence on the contamination point $\bm{x}_t$ is governed by the same function 
$\bm{\psi}_{\beta,\lambda}(\bm{x}_t;\bm{\theta}_g)$ in all cases. 
Hence, the robustness properties of rSDNet under input noise are determined solely
by the behavior of $\bm{\psi}_{\beta,\lambda}$ with respect to $\bm{x}_t$.
\qed
\end{remark}

Now, to study the nature of $\bm{\psi}_{\beta, \lambda}(\bm{x}_t, \bm{\theta}_g)$, recall that the (model) class probabilities 
arise from an NN with softmax output layer as specified in (\ref{EQ:NN-model}). Its (sub-)gradient has the form
\begin{equation*}
\nabla_{\bm{\theta}} p_j(\bm{x}; \bm{\theta}) = p_j(\bm{x}; \bm{\theta}) \left[\nabla_{\bm{\theta}}z_j(\bm{x}_t; \bm{\theta}) 
- \sum_{k=1}^J p_j(\bm{x}; \bm{\theta})\nabla_{\bm{\theta}} z_k(\bm{x}_t; \bm{\theta})\right],	
~~~j=1, \ldots, J.
%	= \frac{\left(\sum_{k=1}^J e^{z_k(\bm{x}_t; \bm{\theta})}\right)e^{z_j(\bm{x}_t;\bm{\theta})} \nabla_{\bm{\theta}}z_j(\bm{x}_t; \bm{\theta}) - e^{z_j(\bm{x}_t;\bm{\theta})}\left(\sum_{k=1}^J e^{z_k(\bm{x}_t; \bm{\theta})}\nabla_{\bm{\theta}} z_j(\bm{x}_t; \bm{\theta})\right)}{\left(\sum_{k=1}^J e^{z_k(\bm{x}_t; \bm{\theta})}\right)^2},
\end{equation*}
Because $0\le p_j,p_j^*\le1$, the weights $u_j(\bm{x};\bm{\theta})$ in \eqref{EQ:psi} are bounded for all $j\geq 1$.
Consequently, the boundedness of $\bm{\psi}_{\beta,\lambda}(\bm{x}_t;\bm{\theta}_g)$, and hence that of the IF, 
depends primarily on the growth of the network (sub-)gradients $\nabla_{\bm{\theta}}p_j(\bm{x}_t;\bm{\theta})$, 
which may be unbounded depending on the architecture.
The IF remains bounded over an unbounded feature space if either $\bm{p}(\cdot; \bm{\theta}_g) \equiv \bm{p}^*$,
or   $\nabla_{\bm{\theta}}p_j(\bm{x}_t; \bm{\theta})$ is bounded in $\bm{x}_t$. 
The first condition corresponds to a correctly specified conditional model, in which case the IF becomes identically zero, 
although this scenario rarely holds in practice even with highly expressive NNs. 
Thus, local robustness of rSDNet under feature contamination is ensured 
when the network gradients with respect to parameters remain controlled.
For any given architecture, the parameters $\beta>0$ and $\lambda<0$ further mitigate the growth of $\bm{\psi}_{\beta,\lambda}$ 
through the down-weighting functions $u_j(\bm{x};\bm{\theta})$, $j=1, \ldots, J$. 
Increasing $\beta$ and decreasing $\lambda$ strengthen this attenuation, 
thereby improving tolerance to moderate feature contamination. 
Such effects are further clarified though the following simple example.

\begin{figure}[H]
	\centering
	\begin{subfigure}[b]{0.48\textwidth}
		\centering
		\includegraphics[width=\textwidth]{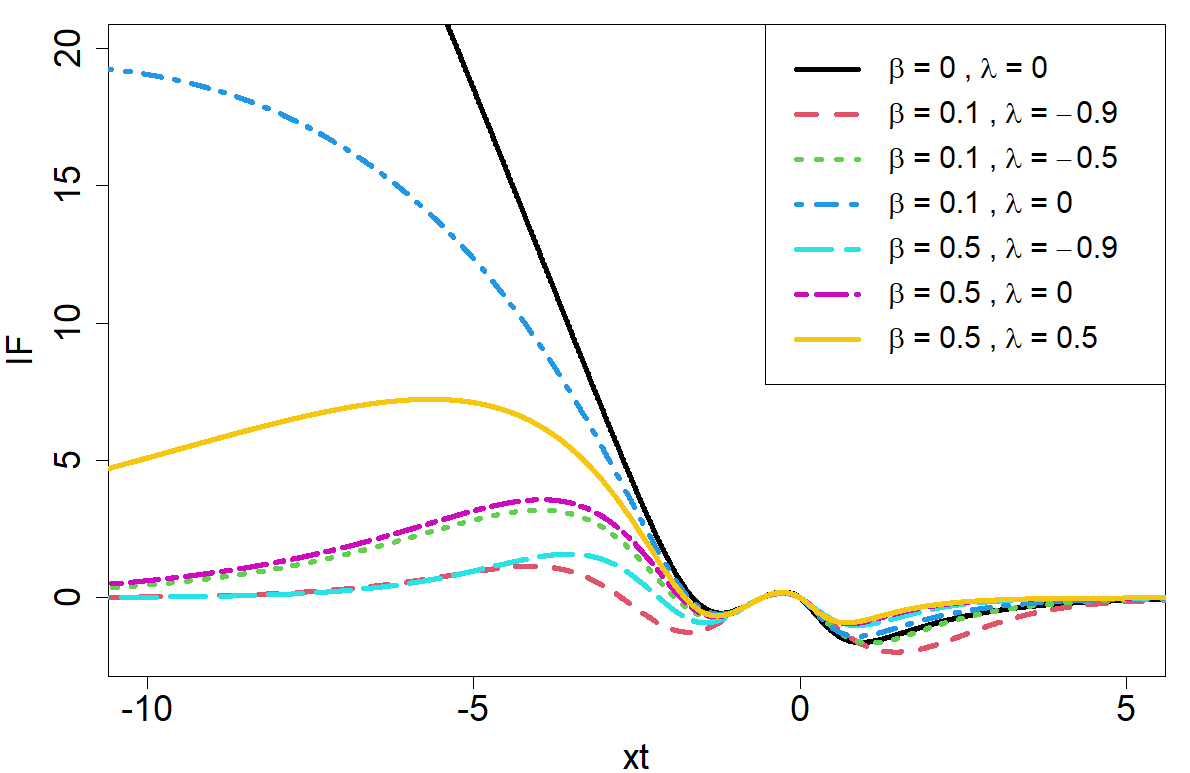}
		\caption{IFs for $\theta_1$}
		\label{fig:M1_theta1}
	\end{subfigure}
	%     \hfill
	\begin{subfigure}[b]{0.48\textwidth}
		\centering
		\includegraphics[width=\textwidth]{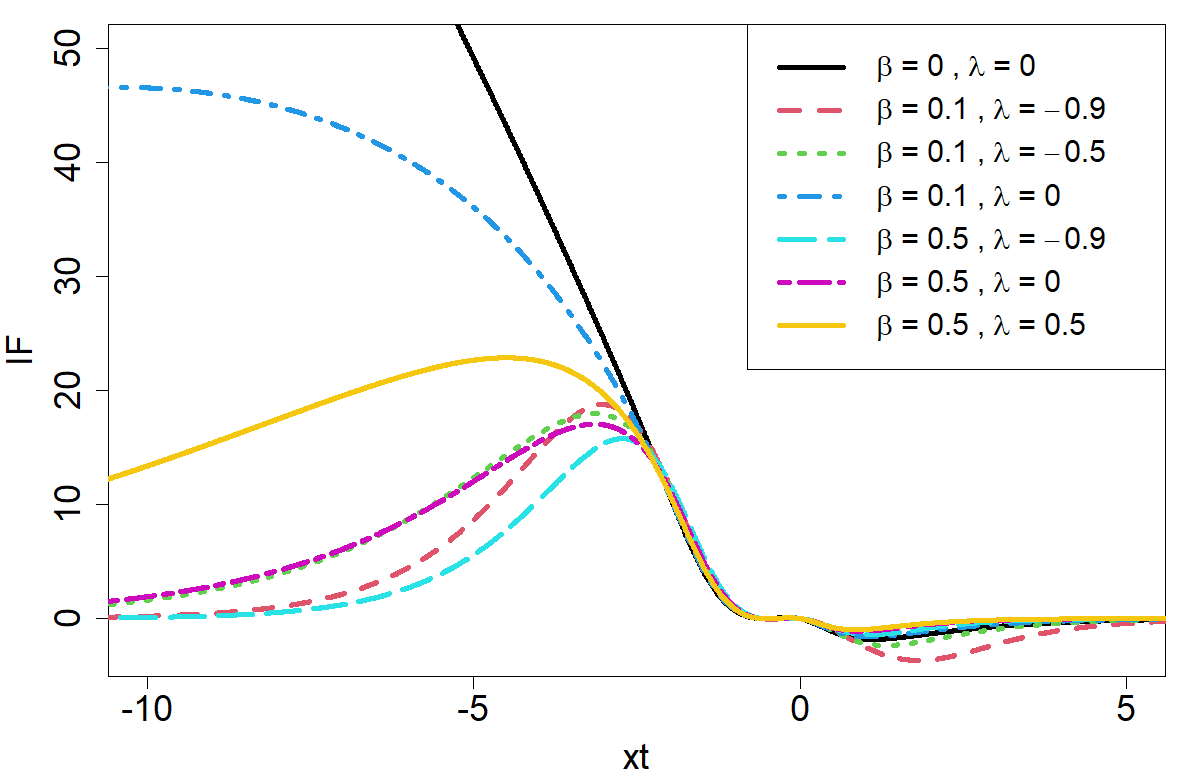}
		\caption{IFs for $\theta_2$}
		\label{fig:M1_theta2}
	\end{subfigure}
	\caption{IFs for the MSDFs of parameters under the NN model (M1) in Example \ref{EX:IF_ex},
		for different tuning parameters $(\beta, \lambda)$.}
	\label{fig:IF_M1}
\end{figure}

\begin{figure}[H]
	\centering
	\begin{subfigure}[b]{0.48\textwidth}
		\centering
		\includegraphics[width=\textwidth]{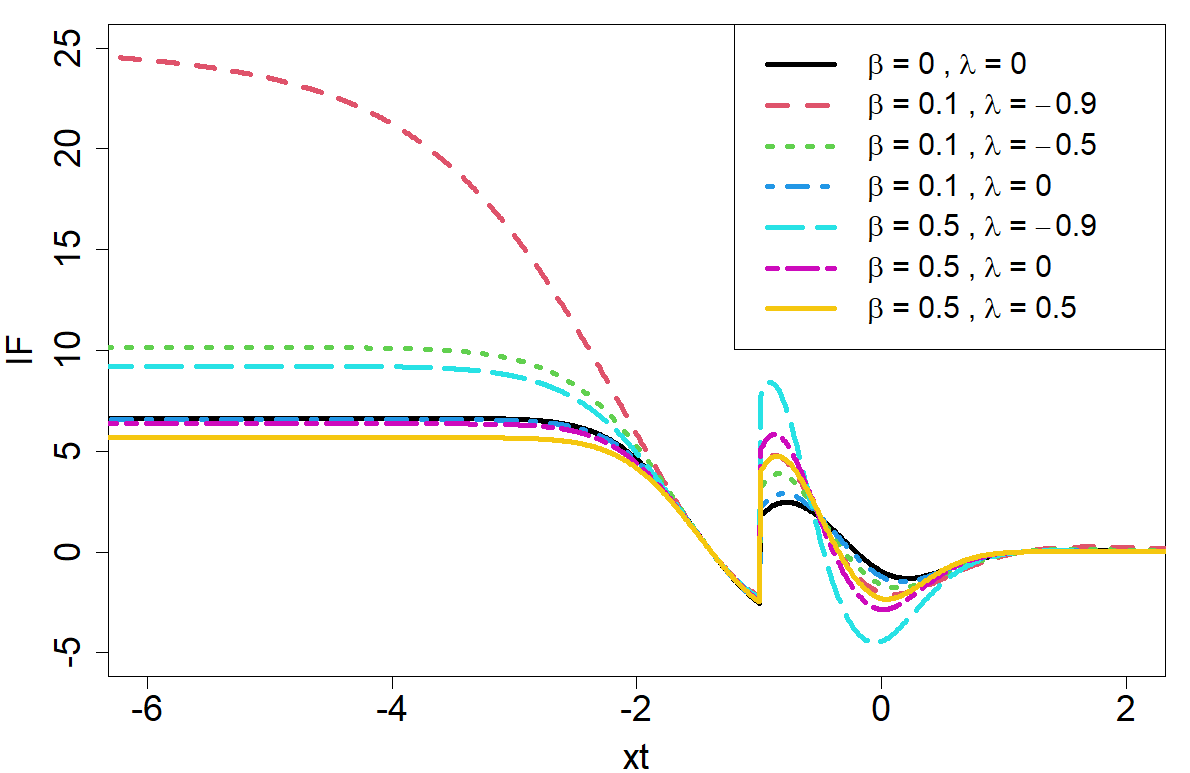}
		\caption{IFs for $\theta_1$}
		\label{fig:M2_theta1}
	\end{subfigure}
	%     \hfill
	\begin{subfigure}[b]{0.48\textwidth}
		\centering
		\includegraphics[width=\textwidth]{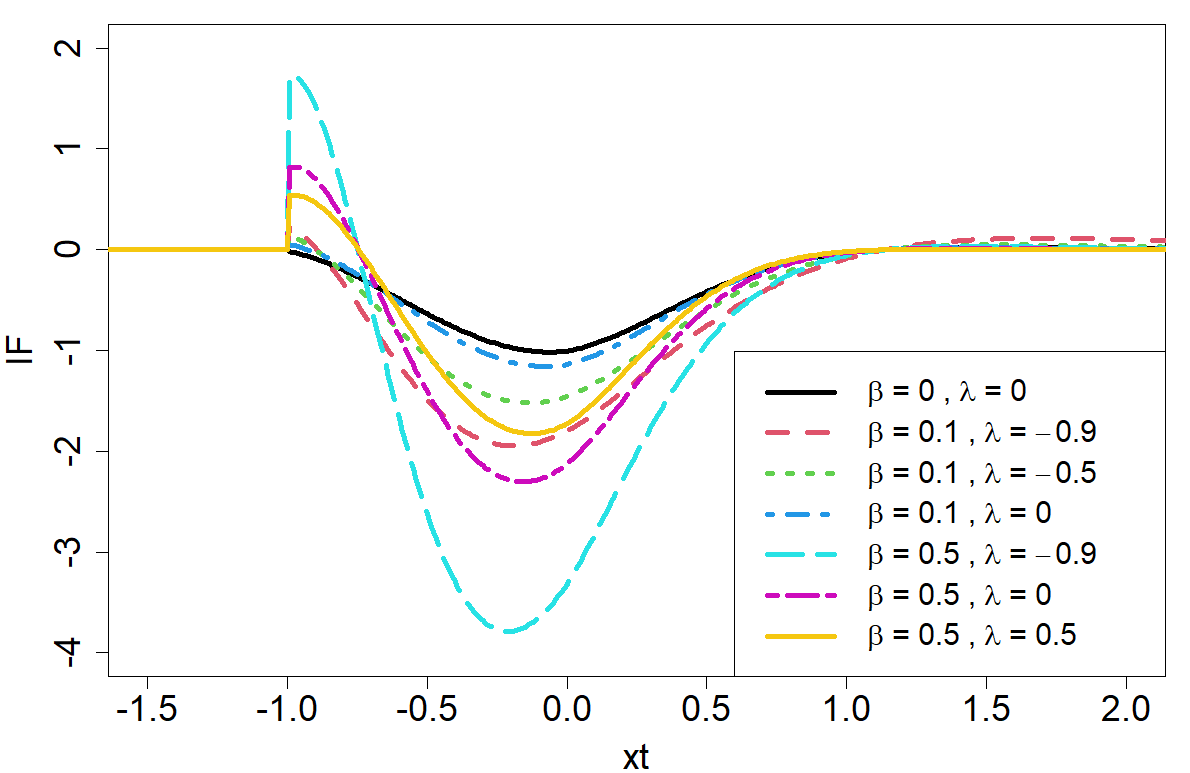}
		\caption{IFs for $\theta_2$}
		\label{fig:M2_theta2}
	\end{subfigure}
	\begin{subfigure}[b]{0.48\textwidth}
		\centering
		\includegraphics[width=\textwidth]{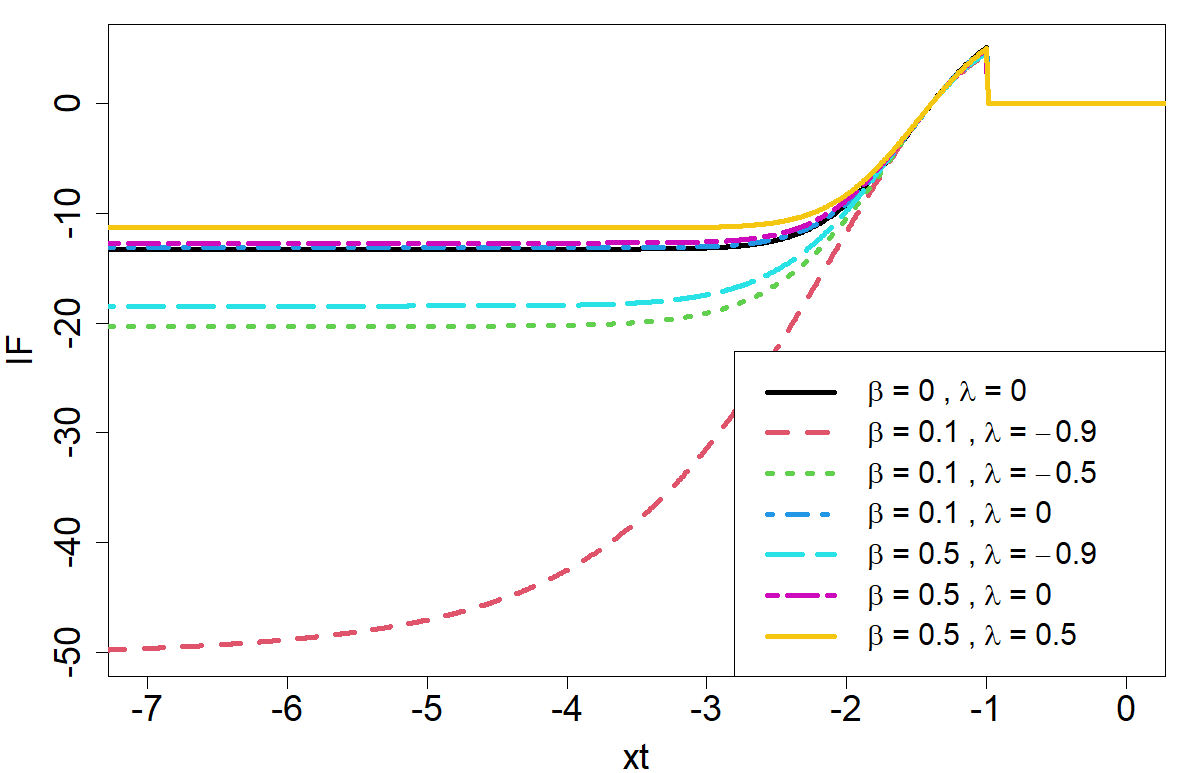}
		\caption{IFs for $\theta_5$}
		\label{fig:M2_theta5}
	\end{subfigure}
	\begin{subfigure}[b]{0.48\textwidth}
		\centering
		\includegraphics[width=\textwidth]{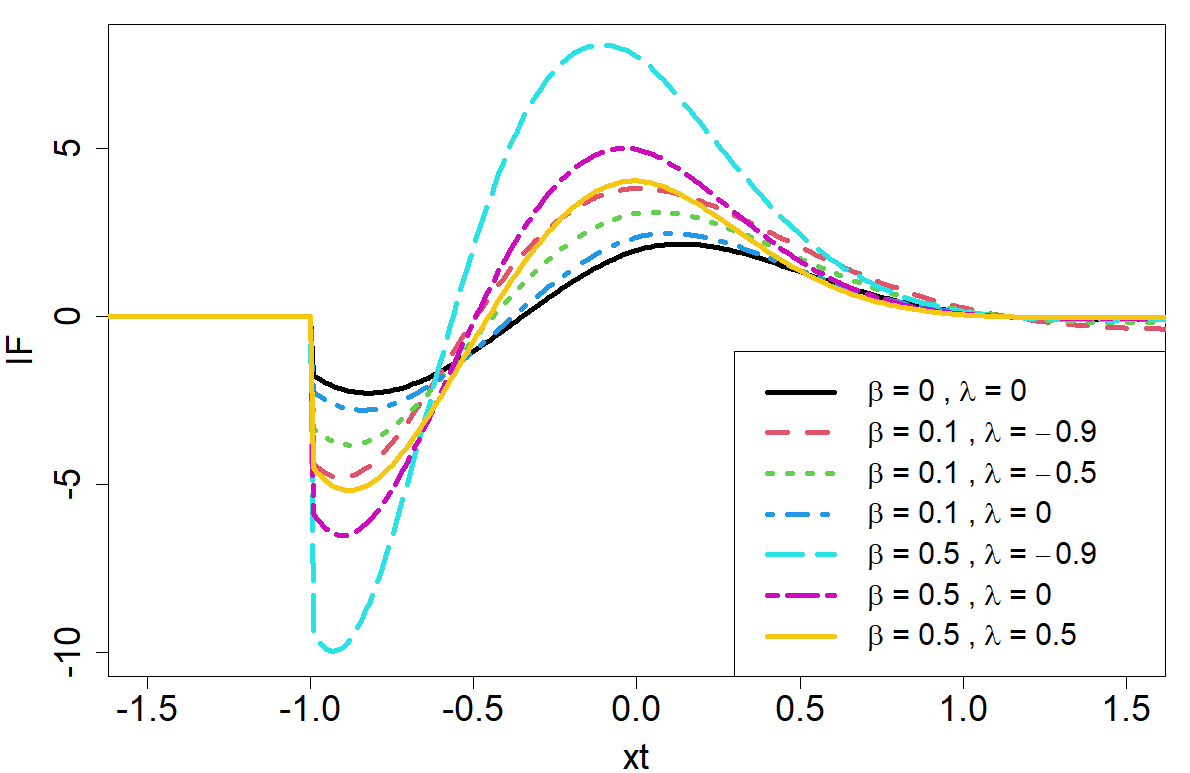}
		\caption{IFs for $\theta_6$}
		\label{fig:M2_theta6}
	\end{subfigure}
	\caption{IFs for the MSDFs of parameters under the NN model (M2) in Example \ref{EX:IF_ex},
		for different tuning parameters $(\beta, \lambda)$.}
	\label{fig:IF_M2}
\end{figure}

\begin{figure}[H]
	\centering
	\begin{subfigure}[b]{0.48\textwidth}
		\centering
		\includegraphics[width=\textwidth]{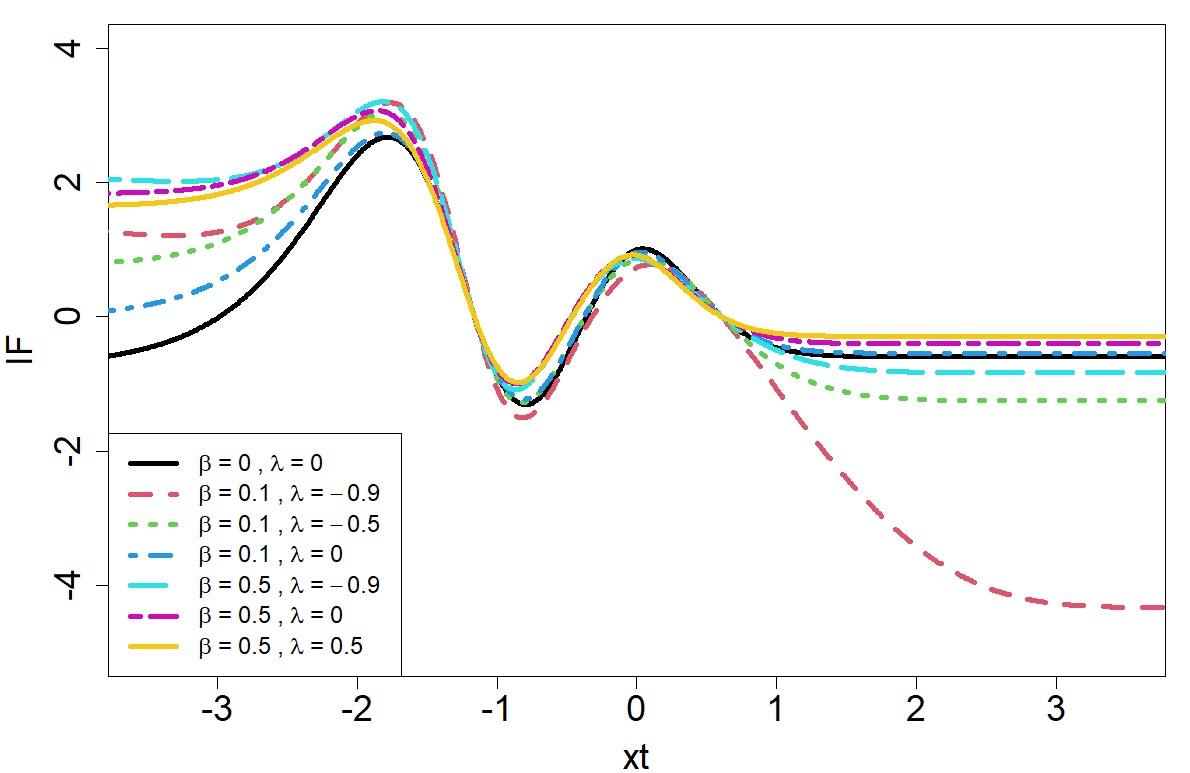}
		\caption{IFs for $\theta_1$}
		\label{fig:M3_theta1}
	\end{subfigure}
	%     \hfill
	\begin{subfigure}[b]{0.48\textwidth}
		\centering
		\includegraphics[width=\textwidth]{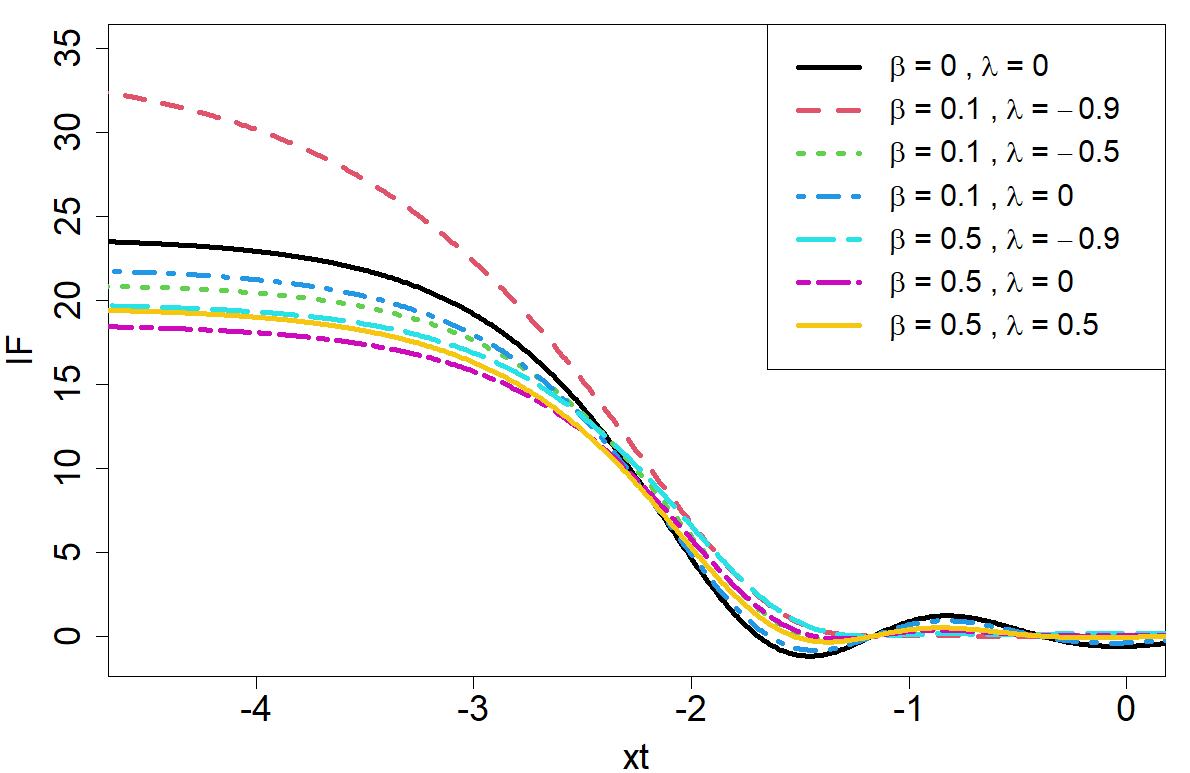}
		\caption{IFs for $\theta_2$}
		\label{fig:M3_theta2}
	\end{subfigure}
	\begin{subfigure}[b]{0.48\textwidth}
		\centering
		\includegraphics[width=\textwidth]{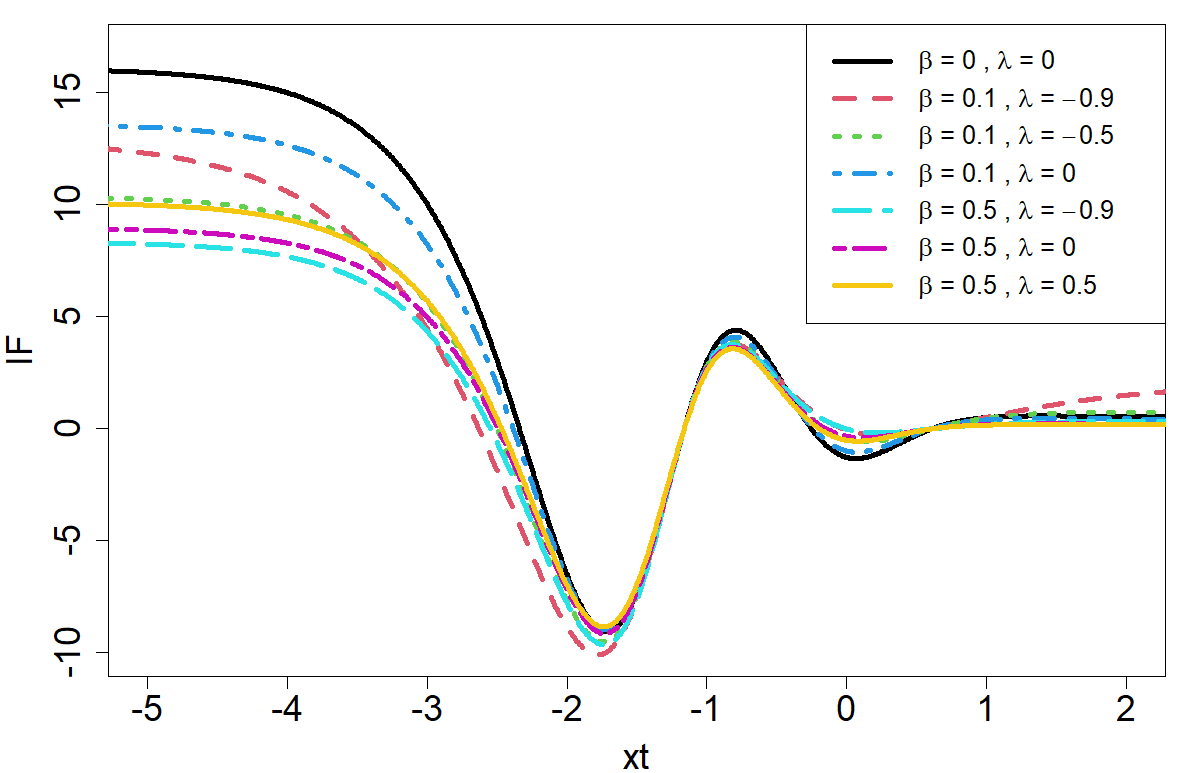}
		\caption{IFs for $\theta_5$}
		\label{fig:M3_theta5}
	\end{subfigure}
	\begin{subfigure}[b]{0.48\textwidth}
		\centering
		\includegraphics[width=\textwidth]{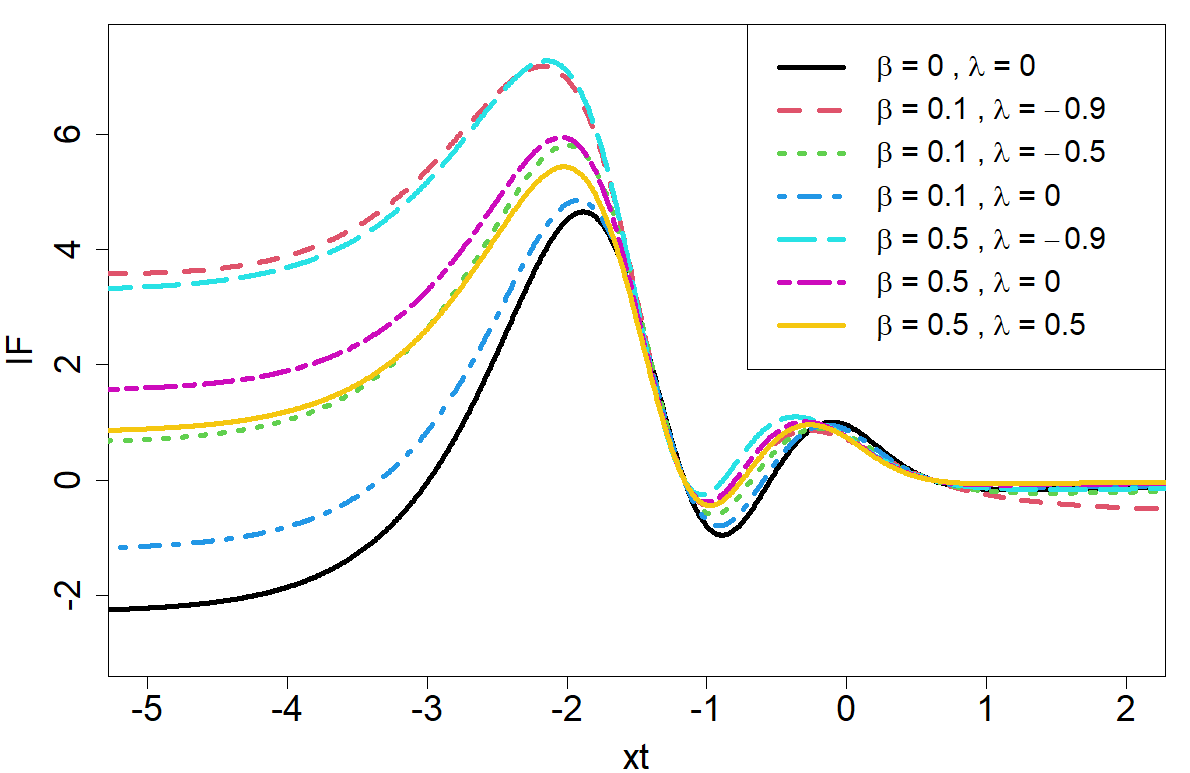}
		\caption{IFs for $\theta_6$}
		\label{fig:M3_theta6}
	\end{subfigure}
	\caption{IFs for the MSDFs of parameters under the NN model (M3) in Example \ref{EX:IF_ex},
		for different tuning parameters $(\beta, \lambda)$.}
	\label{fig:IF_M3}
\end{figure}

\begin{example}\label{EX:IF_ex}
Consider the binary classification problem with a single input feature $x\sim N(0,1)$,  $\mathcal{X}=\mathbb{R}$,
and true posterior class probability $\bm{p}^*(x) = (p_1^*(x), 1 - p_1^*(x))$, where 
$$
p_1^*(x) = \frac{e^{\varkappa(x)}}{1+e^{\varkappa(x)}}, ~~~ 
\varkappa(x) = sin(x) + e^x + x^{5/3}.
$$
We model it using neural classifiers (\ref{EQ:NN-model}) with $J=2$, $p_2(x, \bm{\theta}) = 1- p_1(x, \bm{\theta})$,  
and $z_1(x, \bm{\theta})$ in the definition of $p_1(x, \bm{\theta})$ being specified by three NN architectures as follows: 
\begin{itemize}
\item[(M1)] Single layer perceptron: $z_1(x;\bm{\theta}) = \theta_1 + \theta_2 x$, with $\bm\theta = (\theta_1, \theta_2)^\top$.
\item[(M2)] ReLU hidden layer:  $z_1(x;\bm{\theta}) = \theta_5 + \theta_6 \varphi(\theta_1+\theta_2 x) 
+ \theta_7 \varphi(\theta_3+\theta_4 x)$, with $\varphi(s) = \max(0, s)$ and $\bm\theta = (\theta_1, \theta_2, \ldots, \theta_7)^\top$.
\item[(M3)] $\tanh$ hidden layer:  $z_1(x;\bm{\theta}) = \theta_5 + \theta_6 \tanh(\theta_1+\theta_2 x) + \theta_7 \tanh(\theta_3+\theta_4 x)$, with $\bm\theta = (\theta_1, \theta_2, \ldots, \theta_7)^\top$.
\end{itemize}
It is straightforward to verify that the (sub-)gradient of $z_1(x;\bm{\theta})$ is bounded in $x\in\mathbb{R}$ only for (M3),
while it grows linearly in $x$ for both (M1) and (M2). 

To examine local robustness of rSDNet across the three architectures, we numerically evaluate the IFs of the MSDF for $\bm{\theta}$
at $\bm{\theta}_g = (1, \ldots, 1)^T$. The expectations $E_{G_X}$, appearing in the expression of IF (Theorem \ref{thm:IF_rSDNet}) 
are approximated by the empirical average based on a random sample of size $n=100$ drawn from $N(0,1)$. 
The resulting IFs are plotted in Figures \ref{fig:IF_M1}--\ref{fig:IF_M3}.  
For brevity, under (M2) and (M3), we present the IFs only for $\theta_1, \theta_2, \theta_5$ and $\theta_6$, 
since the remaining parameters exhibit similar behavior.
It is evident from the figures that, for suitably chosen tuning parameters $(\beta, \lambda)$, 
the IFs remain well-controlled, often bounded, with respect to the contamination point $x_t$,
thereby indicating the local robustness of the corresponding rSDNet. 
\qed 
\end{example}

%Neural networks trained using the standard cross-entropy loss satisfy estimating equations similar to those above but with
%\[
%u_j(\bm{x};\bm{\theta}) = p_j(\bm{x};\bm{\theta}) - p_j^*(\bm{x}).
%\]
%Unlike rSDNet, cross-entropy does not involve the stabilizing power terms associated with $\beta$ and $B$. As a result, the contribution of observations with extreme probability values may grow more rapidly in the estimating equations. Consequently, while both approaches may exhibit unbounded influence functions under severe covariate contamination due to network gradients, rSDNet provides additional robustness through bounded loss contributions and reduced sensitivity to label-related perturbations.
%
%The influence function analysis indicates that the robustness of rSDNet to feature contamination depends on two main factors: (i) the robustness properties of the SD-loss itself, and (ii) the behavior of the neural network gradients with respect to the parameters. While the SD-loss controls the contribution of the response component, extreme covariate values may still induce large gradients in the network parameters. Therefore, the practical robustness of rSDNet to feature contamination depends critically on architectural choices and gradient behavior.

\section{Empirical Evaluation on Image Classification}\label{SEC:numericals}

\subsection{Experimental setups: Datasets and NN architectures}
\label{SEC:exp_setup}
To assess the finite-sample performance of the proposed rSDNet, 
we conducted experiments on the following three benchmark image classification datasets,
which were selected because of their widespread adoption in the NN literature and varying levels of classification difficulty.
%available in the \texttt{TensorFlow} {or any other standard library used for NN training}
\begin{itemize}
    \item \textbf{MNIST dataset:}
    It consists of $70000$ grayscale images of handwritten digits (0--9), each with a spatial resolution of $28\times28$ pixels. 
    Considering each pixel intensity as an input feature, we have $784$ covariates per image,
    which were linearly normalized, from their original range [0, 255] to the interval $[0,1]$, prior to model training.
    
    To build a neural classifier for this dataset, we employed a fully connected multilayer perceptron (MLP) 
    consisting of two hidden layers with 128 neurons each and ReLU activation functions. 
    The output layer contained 10 neurons with a softmax activation function to model the categorical class labels 0--9.

    \item \textbf{Fashion-MNIST dataset:} 
    It also contains $70000$ grayscale images of size 28 × 28 pixels, 
    categorized into 10 clothing classes (e.g., T-shirt/top, trouser, sandal, coat, etc.). 
    As in the MNIST dataset, each image corresponds to 784 input features representing pixel intensities,
    rescaled to [0, 1] before training the NN model.
 
 	The NN architecture was taken to be a fully connected MLP with two hidden layers 
 	containing 200 and 100 neurons, respectively, both using ReLU activation functions,
 	and 10 softmax neurons in the output layer.
%    \[
%    \begin{aligned}
%    28 \times 28 \times 3 &\mbox{ (input image)}
%    \;\rightarrow\; \text{Dense}(200,\, \text{ReLU})\\
%    & \;\rightarrow\; \text{Dense}(100,\, \text{ReLU}) 
%      \;\rightarrow\; \text{Dense}(10,\, \text{Softmax}) \mbox{ (Output)}
%    \end{aligned}
%    \]

    \item \textbf{CIFAR-10 dataset:}
    It consists of 60000 colored (RGB) images representing 10 object categories. 
    Each image has a resolution of 32 × 32 pixels with three color channels, yielding 32 × 32 × 3 = 3072 input features,
    which were normalized to [0, 1] as before.
%    Pixel intensities were normalized from the range [0, 255] to [0, 1] for these data as well. 
    
    Given the increased complexity, here we adopted a convolutional NN (CNN) architecture,
    consisting of two convolutional layers with 32 and 64 filters (kernel size 3 × 3, ReLU activation), 
    each followed by 2 × 2 max-pooling layers. 
    The convolutional feature maps were flattened and passed through a fully connected layer with 512 ReLU units, 
    followed by a 10-unit softmax output layer as follows:
    \[
    \begin{aligned}
    32 \times 32 \times 3 &\mbox{ (input image)}
    \;\rightarrow\; \text{Conv2D}(32,\, 3\times 3,\, \text{ReLU}) 
       \;\rightarrow\; \text{MaxPool}(2\times 2) \\
    &\;\rightarrow\; \text{Conv2D}(64,\, 3\times 3,\, \text{ReLU}) 
       \;\rightarrow\; \text{MaxPool}(2\times 2) \;\rightarrow\; \text{Flatten} \\
    & \;\rightarrow\; \text{Dense}(512,\, \text{ReLU}) 
       \;\rightarrow\; \text{Dense}(10,\, \text{Softmax}) \mbox{ (Output)}
    \end{aligned}
    \]
\end{itemize}

\noindent
\textbf{Generic Model Training and Evaluation Procedure:}\\
Although all three datasets provide default training–test splits, available in the \texttt{TensorFlow} library of Python, 
we combined the original training and test partitions, and randomly generated $k>1$ folds to evaluate 
the performance of the trained neural classifiers using $k$-fold cross validation. 
For each dataset, we applied the proposed rSDNet with various tuning parameter combinations (and 250 epochs),
and compared the results with existing benchmark loss functions/classifiers such as CCE, MAE, TCCE($\delta$), rKLD, 
SCE($\alpha, \beta$), GCE($q$), and FCL($\mu$). Here $\delta$ denotes the trimming proportion, 
and $(\alpha, \beta)$, $q$ and $\mu$ are tuning parameters defining the respective loss functions. 
In all cases, model training was performed on both clean and suitably contaminated versions of the training folds.  
Final performance of each classifier was evaluated using average $k$-fold cross-validated classification accuracy 
computed on uncontaminated test/validation folds. 
The number of folds was set to $k=7$ for MNIST and Fashion-MNIST, and $k=6$ for CIFAR-10 to match their original training-test split ratio. 
%{The 70000 images of the MNIST data set come with a default train-test split ratio of 60000:10000. So, to match this ratio, we took $k=7$ for MNIST data. The reason for taking $k=7$ for the Fashion-MNIST data is also the same. Further, we took $k=6$ in case of the CIFAR-10 data to match the original train-test split ratio of 50000:10000.}
%For rSDNet, each NN model was trained for 250 epochs across all datasets.

\subsection{Performances under clean data} 
\label{SEC:emp_CLD-perform}
The average cross-validated (CV) classification accuracies of all NN models trained on the non-contaminated (clean) datasets are reported in 
Table \ref{tab:no-cont-CV}, for the proposed rSDNet and existing benchmark learning algorithms.
In such cases of clean data, the standard CCE loss achieves strong performance across all three datasets, 
with respective accuracy of 0.9816, 0.8891, and 0.6782. 
The SCE loss and GCE with  $q=0.5$ show comparable results, particularly on MNIST and Fashion-MNIST datasets.
Our rSDNet consistently matches, and in several configurations slightly improves upon, the CCE baseline. 
For MNIST, multiple rSDNet parameter combinations achieve accuracies slightly above 0.980, 
with the best performance of 0.9809 attained at $(\beta,\lambda)=(0.3,0)$, which is essentially identical to CCE. 
On Fashion-MNIST, rSDNet reaches a maximum accuracy of 0.8948 at $(\beta,\lambda)=(0.1, -0.5)$, 
slightly outperforming  CCE (0.8891) and SCE (0.8911). For CIFAR-10, the best rSDNet accuracy is 0.6729 at $(\beta,\lambda)=(0.1,0)$, 
which is again very competitive with CCE (0.6782) and SCE (0.6785) losses. 
Overall, rSDNet maintains competitive performance across a broad range of  $(\beta,\lambda)$ choices, 
demonstrating no significant loss in efficiency under clean conditions.

In contrast, several existing robust alternatives exhibit clear degradation in accuracy when no contamination is present. 
TCCE shows progressively worse performance as the trimming proportion increases, 
which is expected as trimming removes informative observations in the absence of outliers. 
%While TCCE(0.1) remains reasonably competitive, larger trimming levels severely reduce accuracy on MNIST and Fashion-MNIST. 
%For example, TCCE(0.3) drops to 0.7787 on MNIST and below 0.30 on Fashion-MNIST. 
The MAE loss also performs adequately on MNIST (0.9770) but substantially underperforms on Fashion-MNIST (0.7940) 
and completely fails on CIFAR-10 (0.1004) datasets, indicating its optimization instability in complex scenarios. 
A similar phenomenon appears for FCL at certain parameter settings. 
In particular, FCL with $\mu=0$ collapses to near-random performance (accuracy 0.1) on CIFAR-10, 
and larger values $\mu \ge 0.5$ lead to near-random accuracy on both MNIST and Fashion-MNIST.
GCE with $q=0.7$ also shows noticeable deterioration on CIFAR-10 (0.5590), although it remains competitive on MNIST.

These results shows extremely high efficiency of the proposed rSDNet 
compared to its existing robust competitors when trained on clean data.
Importantly, this stability holds across a wide range of tuning parameter values, 
suggesting that rSDNet (unlike other robust learning algorithms) does not incur 
a performance penalty in the absence of contamination.

\subsection{Robustness against uniform label noises}
\label{SEC:exp_ULN}

In order to illustrate the performances under label corruption, uniform label noise was introduced 
by randomly replacing the true class label of a specified proportion ($\eta$) of training observations 
with one of the remaining class labels. Models were trained on such contaminated training folds, 
while performances were evaluated on clean test folds as before.
The resulting average cross-validated accuracies for contamination level $\eta=$ 0.1 to 0.5 
are presented in Tables \ref{tab:mnist-cv}--\ref{tab:cifar10-cv}, respectively, for the three datasets. 

For all datasets, the CCE loss shows a steady and substantial performance degradation as noise increases.
Similar deterioration is observed also for SCE and FCL with small $\mu \leq 0.25$.
MAE and GCE($q=0.7$) remain highly stable for MNIST data, maintaining accuracy above 0.95 even at $\eta=0.5$.
MAE also shows moderate robustness for Fashion-MNIST data but fluctuates and falls below stronger competitors at higher contamination levels;
it however collapses to near-random performance ($\approx$ 0.10) across all contamination levels for the more complex CIFAR-10 dataset.
Among existing robust losses, GCE performs well for Fashion-MNIST data, maintaining relatively high accuracy (0.8339) even at $\eta=0.5$ for $q=0.7$,
while TCCE(0.2) achieves the highest accuracy for CIFAR-10 data at low contamination. 
%However, its performance also deteriorates substantially as noise increases.
%The rKLD configuration $(0,-1)$, however, performs substantially worse than other rSDNet variants, 
%particularly at higher contamination levels.

In contrast, the proposed rSDNet consistently exhibits strong robustness for all three datasets 
for small $\beta>0$ (0.05--0.1) and  $\lambda<0$ (between $-0.5$ and $-1$). 
%It thus maintains clear advantages over CCE and most competing losses when contamination is moderate to high.
%However, as $\beta$ increases (e.g., $\beta \geq 0.3$)  or as $\lambda$ moves toward zero or positive values, 
%robustness deteriorates noticeably. 
While some existing losses provide robustness under specific settings (e.g., MAE on MNIST, or GCE with $q=0.7$ for Fashion-MNIST), 
their performance is often dataset-dependent or unstable for more complex data (e.g., CIFAR-10). 
The proposed rSDNet, however, provides consistently competitive or superior performance across datasets 
and contamination levels without collapsing even under high label noise.

\subsection{Stability against diverse adversarial attacks}
\label{SEC:exp_AdvAttack}

To further assess the robustness of the proposed rSDNet, 
we conducted the same empirical experiments under four widely studied white-box adversarial attacks, 
namely the fast gradient sign method (FGSM) \citep{goodfellow2014explaining}, 
projected gradient descent (PGD) \citep{madry2017towards}, Carlini–Wagner (CW) attack \citep{carlini2017towards}, 
and DeepFool attack \citep{moosavi2016deepfool}. 
Adversarial examples were generated from the MNIST dataset using a surrogate NN model, 
having a single hidden layer of 64 ReLU nodes and 10 softmax output nodes, trained with the CCE loss for 250 epochs. 
Separate training folds were constructed using fully adversarial images from each attack with suitable hyperparameter values\footnote{
FGSM with perturbation magnitude 0.3; PGD with attack step size 0.01 and maximum perturbation bound 0.3;
untargeted CW  with  learning rate 0.01;  DeepFool  with overshoot parameter 0.02.
Maximum 100 iterations are used in last three cases.}.
The same NN models, as before, were trained based on these training datasets.
In addition to the clean test accuracy, here we also computed average accuracies on adversarially perturbed test folds under the same attack.
The results are reported in Table \ref{tab:adv-attack}.

Under such adversarial training, CCE achieves high adversarial test accuracy (0.9776 -- 0.9949 for FGSM/PGD, and 0.9815 under CW), 
but its clean test accuracy can drop substantially, particularly under PGD (0.6516). 
TCCE shows systematic degradation in clean accuracy as the trimming proportion increases, 
with sharper decline under stronger attacks. For example, under PGD training, 
clean accuracy falls below 0.47 for trimming levels $\geq 0.1$,
indicating that excessive trimming removes substantial useful information even in structured adversarial settings.

In contrast, rSDNet maintains competitive or superior performance across a wide range of tuning parameters. 
Under FGSM training, it achieves the highest clean test accuracy (0.9235 at $(0.9,0.5)$), exceeding CCE, 
while maintaining comparable adversarial accuracy ($\approx 0.975$). 
Under PGD, rSDNet attains the best adversarial accuracy (0.9958 at $(0.3,-1)$) 
and slightly improves clean accuracy over CCE in certain configurations (maximum 0.6582 at $(1,0)$). 
For CW and DeepFool attacks, rSDNet performs similarly to the CCE, 
with adversarial accuracies consistently between 0.978 and 0.985 and clean accuracies close to the best observed values. 
The highest DeepFool adversarial accuracy (0.9854) is achieved at by rSDNet$(0.5,0.5)$.
Generally, rSDNet with moderate-to-large $\beta>0$  combined with $\lambda\in[-1, 0.5]$ yields strong adversarial robustness 
without sacrificing clean performance. No rSDNet configuration exhibits a severe decline in performance,
and the results remain tightly concentrated across parameter choices.
All these validate that rSDNet preserves predictive accuracy under adversarial training
while providing stable and competitive robustness across diverse attack mechanisms, 
often matching or exceeding the CCE baseline and avoiding the degradation observed for heavily trimmed losses.

\subsection{On the choice of rSDNet tuning parameters}

Based on theoretical considerations and empirical evidence, 
it is evident that the practical performance of rSDNet depends heavily on the choice of its tuning parameters $(\beta, \lambda)$.
The pattern is consistent with prior applications of $S$-divergences in robust statistical inference; see, e.g., \cite{ghosh2017generalized,ghosh2015asymptotic,roy2026asymptotic}.
In the present context of neural learning as well, 
rSDNet achieves optimal performance under uncontaminated data when $\beta$ and $|\lambda|$ are small, 
whereas larger values of $\beta>0$ and $\lambda<0$ are required to obtain stable results in the presence of increasing data contamination. 
Since the extent of potential contamination is unknown in most cases, 
we recommend selecting $\beta\in(0, 0.1]$ and $\lambda\in[-1, -0.5]$,
which provides a favorable robustness-efficiency trade-off across all scenarios considered in our empirical studies.  	
In practice, an optimal pair $(\beta,\lambda)$ for a given datasets may be determined 
via cross-validation over a grid of feasible values within these ranges.

\section{Concluding remarks} \label{conclusion}

In this work, we utilized a broad and well-known family of statistical divergences, namely the SD family, 
to construct a flexible class of loss functions for robust NN learning in classification tasks. 
The resulting framework, rSDNet, provides a principled, statistically consistent, Bayes optimal classifier,
whose robustness and efficiency can be explicitly controlled via its divergence parameters.
We theoretically characterized the effect of these tuning parameters on the robustness of rSDNet 
and empirically demonstrated its improved stability across benchmark image-classification datasets.
With appropriately chosen parameters, rSDNet preserves predictive accuracy on clean data 
while offering enhanced resistance to label noise and adversarial perturbations. 
These results highlight the practical viability of divergence-based training as a robust alternative 
to conventional cross-entropy learning based on possibly noisy training data.

Nevertheless, to clearly isolate and understand the effects of divergence-based losses relative to existing methods, 
we deliberately restricted our experiments to relatively simple NN architectures, focusing primarily on image classification. 
A systematic theoretical and empirical investigation of rSDNet in modern large-scale deep learning pipelines,
incorporating complex architectures such as residual networks, transformer-based models and hybrid attention-based architectures, 
remains an important direction for future work.
While we provided initial theoretical insights for rSDNet, 
a comprehensive analysis of its convergence dynamics, generalization guarantees, and adversarial robustness 
in deep, non-convex, and potentially non-smooth network architectures is yet to be developed. 
Furthermore, broadening the application of rSDNet to other data modalities, 
such as text, tabular, graph-structured, and multimodal data, 
would significantly enhance its practical impact, enabling robust learning from large, noisy datasets across diverse scientific and industrial domains.
Most importantly, we hope that the theoretical insights and empirical findings presented here will motivate further research 
on statistically grounded loss functions and contribute to the development of reliable, robust, and trustworthy AI systems.

\appendix
\section{Proof of the results}\label{proofs}

\subsection{Proof of Lemma \ref{lem:sd-loss-bounded}}
\label{App:sd-loss-bounded}
%\begin{proof}
	From \eqref{EQ:SD-loss-pis}, we get 
\begin{eqnarray}
	\sum_{j=1}^J \ell_{\beta,\lambda}(\bm{e}_j, \bm{p}(\bm{x};\bm{\theta}))
	= \displaystyle \frac{J}{A}\sum_{j=1}^J p_j^{1+\beta} - \frac{1+\beta}{AB} \sum_{j=1}^J p_j^B + \frac{J^2}{B}.  
\label{EQ:a0}
\end{eqnarray}
We use the following  bounds for sums of powers of a probability vector $\bm{p} = (p_1,\dots,p_J)\in\Delta_J$:
	\begin{itemize}
		\item If $0<r<1$, then $\sum_{j=1}^J p_j^r \in [1, J^{1-r}]$.
		\item If $r\ge 1$, then $\sum_{j=1}^J p_j^r \in [J^{1-r}, 1]$.
	\end{itemize}
Now, since $(\beta, \lambda)\in\mathcal{T}$, using these results,  we get 
	\[
	\sum_{j=1}^J p_j^{1+\beta} \in [J^{-(\beta)}, 1], ~~~~~ 
	\sum_{j=1}^J p_j^B \in [\min(1, J^{1-B}), \max(1, J^{1-B})].
	\]		
Substituting these bounds into (\ref{EQ:a0}) we get the desired result given in  \eqref{eq:sd-loss-bounds}.
%\end{proof}

\subsection{Proof of Theorem \ref{thm:sd-loss-noise-robust}}
\label{App:sd-loss-noise-robust}

We may note that the first inequality of \eqref{EQ:label_noise_bound} follows directly from the definition of $\bm{\theta}_{\beta,\lambda}^*$. \\
Next, to prove the second inequality, we study the excess risk under label contamination as  
\begin{align*}
 R_{\beta,\lambda}^\eta(\bm{\theta}) &= E_{G_{\bm{X}}}\left[E_{g_\eta}\left[\ell_{\beta,\lambda}(\bm{Y}, \bm{p}(\bm{X};\bm{\theta}))|\bm{X}\right]\right], \\
%    &= E_{G_{\bm{X}},g} \left[(1-\eta) \mathcal{L}_{\beta,\lambda}^{SD}(\bm{p}(\bm{X};\bm{\theta}), \bm{Y}) + \frac{\eta}{J-1} \sum_{j:\bm{e}_j\neq\bm{Y}} \mathcal{L}_{\beta,\lambda}^{SD}(\bm{p}(\bm{X};\bm{\theta}), \bm{e}_j)\right]\\
    &= (1-\eta)R_{\beta,\lambda}(\bm{\theta}) + \frac{\eta}{J-1}  \left[E_{G_{\bm{X}}} \left[\sum_{j=1}^J \ell_{\beta,\lambda}(\bm{e}_j, \bm{p}(\bm{X};\bm{\theta}))\right] - R_{\beta,\lambda}(\bm{\theta})\right]\\
    &= \left(1-\frac{J\eta}{J-1}\right) R_{\beta,\lambda}(\bm{\theta}) + \frac{\eta}{J-1} T_{\beta, \lambda}(\bm{\theta}),
\end{align*}
where $T_{\beta, \lambda}(\bm{\theta}) = E_{G_{\bm{X}}} \left[\sum_{j=1}^J \ell_{\beta,\lambda}(\bm{e}_j, \bm{p}(\bm{X};\bm{\theta}))\right]$. 
But, since $\bm{\theta}_{\beta,\lambda}^\eta$ minimizes $R_{\beta,\lambda}^\eta(\bm{\theta})$, we get 
$$
R_{\beta,\lambda}^\eta(\bm{\theta}_{\beta,\lambda}^\eta) \leq R_{\beta,\lambda}^\eta(\bm{\theta}_{\beta,\lambda}^*), 
$$
and hence 
\begin{equation*}
 \left(1-\frac{J\eta}{J-1}\right) R_{\beta,\lambda}(\bm{\theta}_{\beta,\lambda}^\eta) + \frac{\eta}{J-1} T_{\beta, \lambda}(\bm{\theta}_{\beta,\lambda}^\eta)
 \leq  \left(1-\frac{J\eta}{J-1}\right) R_{\beta,\lambda}(\bm{\theta}_{\beta,\lambda}^*) + \frac{\eta}{J-1} T_{\beta, \lambda}(\bm{\theta}_{\beta,\lambda}^*).
\end{equation*}
Rearranging the above, we get 
\begin{equation}
 R_{\beta,\lambda}(\bm{\theta}_{\beta,\lambda}^\eta)  -  R_{\beta,\lambda}(\bm{\theta}_{\beta,\lambda}^*) 
\leq   \frac{\eta}{J-1 - J\eta} \left[T_{\beta, \lambda}(\bm{\theta}_{\beta,\lambda}^*) - T_{\beta, \lambda}(\bm{\theta}_{\beta,\lambda}^\eta)\right].
\label{EQ:a1}
\end{equation}

\noindent
Now, from Lemma~\ref{lem:sd-loss-bounded}, the total SD-loss $T_{\beta, \lambda}(\bm{\theta})$ is bounded for all $\bm{x}$, $\bm{\theta}$,
and thus we get 
$$
T_{\beta, \lambda}(\bm{\theta}_{\beta,\lambda}^*) - T_{\beta, \lambda}(\bm{\theta}_{\beta,\lambda}^\eta)
\leq \max_{\bm{\theta}} T_{\beta, \lambda}(\bm{\theta}) - \min_{\bm{\theta}} T_{\beta, \lambda}(\bm{\theta})
= \frac{1}{A}\left( J - J^{1-\beta} + \frac{1+\beta}{B}|1 - J^{1-B}|\right).
$$  
Substituting it in (\ref{EQ:a1}), we get the desired bound on the excess loss as given in (\ref{EQ:label_noise_bound}).

\subsection{Proof of Theorem \ref{thm:IF_rSDNet}}
\label{APP:IF_rSDNet}

From the definition of the statistical functional $\bm{T}_{\beta,\lambda}$ given in \eqref{EQ:rsdnet-func},
%associated with the estimated model weights through rSDNet, 
we can re-express $\bm{\theta}_g = \bm{T}_{\beta,\lambda}(G_{\bm{X}},g)$ as a solution to the estimating equation
\begin{equation}\label{EQ:a00}
E_{G_{\bm{X}}}\left[\bm{\psi}_{\beta, \lambda}(\bm{X}, \bm{\theta}_g)\right] = \bm{0}.
\end{equation}
Accordingly, $\bm{\theta}_\epsilon = \bm{T}_{\beta,\lambda}(G_{\bm{X}, \epsilon},g)$ satisfies the corresponding estimating equation given by
\begin{equation*}
	E_{G_{\bm{X}}, \epsilon}\left[\bm{\psi}_{\beta, \lambda}(\bm{X}, \bm{\theta}_\epsilon)\right] = \bm{0},
\end{equation*}
which can be expanded to the form:
\begin{equation} \label{IF-st1}
	(1-\epsilon) E_{G_{\bm{X}}}\left[\bm{\psi}_{\beta, \lambda}(\bm{X}, \bm{\theta}_\epsilon) \right] +
	 \epsilon \bm{\psi}_{\beta, \lambda}(\bm{x}_t, \bm{\theta}_\epsilon) = \bm{0}.
\end{equation}

\noindent
Now, differentiating both side of \eqref{IF-st1} with respect to $\epsilon$, and evaluating at $\epsilon=0$ using (\ref{EQ:a00}), we get
\begin{align} \label{IF-est-eqn-1}
	\nonumber
E_{G_{\bm{X}}}\left[\nabla_{\bm{\theta}}\bm{\psi}_{\beta, \lambda}(\bm{X}, \bm{\theta}_g) \right] 
\mathcal{IF}(\bm{x}_t, \bm{T}_{\beta,\lambda}, (G_{\bm{X}}, g)) = \bm{\psi}_{\beta, \lambda}(\bm{x}_t, \bm{\theta}_g).
\end{align}
The expression \eqref{EQ:IF-final} of the IF is then obtained by solving  the above equation via standard theory of linear equations.

\section{Convergence of rSDNet: An empirical study}
\label{APP:conv}

\begin{figure}[H]
	\centering
	\begin{subfigure}[b]{0.7\textwidth}
		\centering
		\includegraphics[width=1\linewidth]{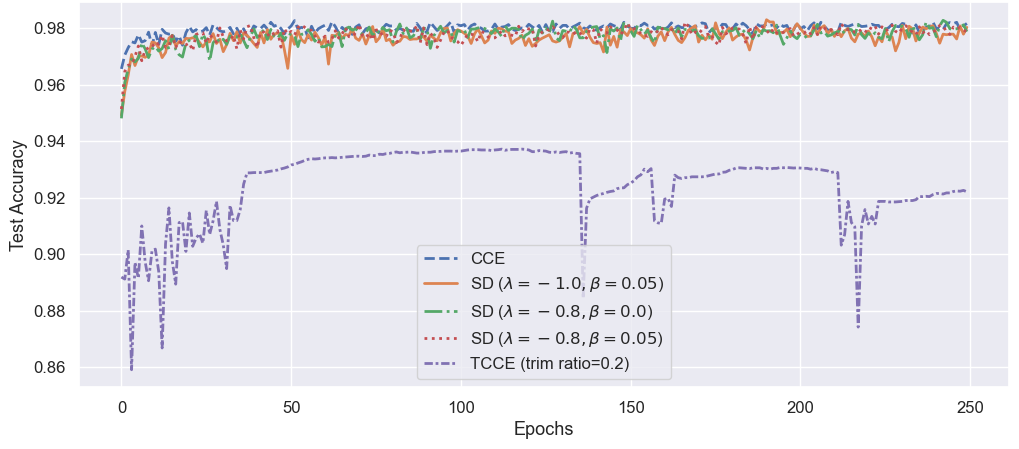}
		\caption{Clean training data}
		\label{fig:mnist-history_plot_0}
	\end{subfigure}
	\hfill
	\begin{subfigure}[b]{0.7\textwidth}
		\centering
		\includegraphics[width=1\linewidth]{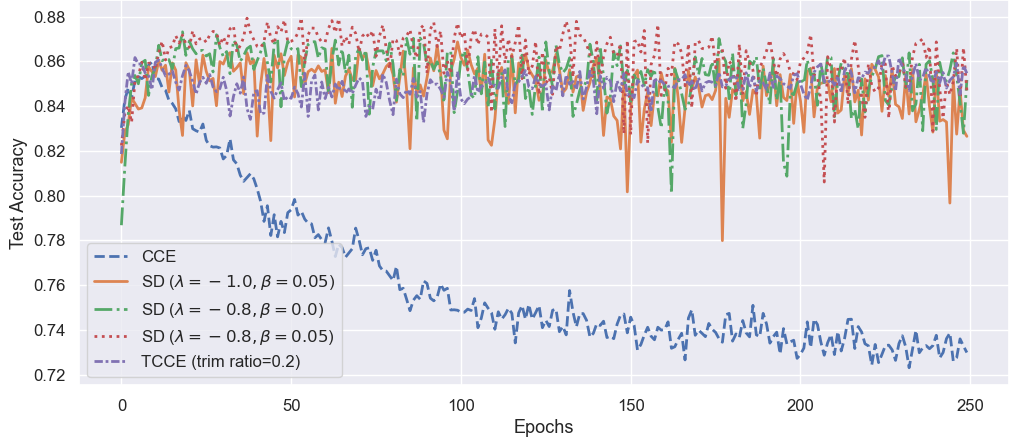}
		\caption{Contaminated training data with 20\% uniform label noise}
		\label{fig:mnist-history_plot_20}
	\end{subfigure}
	\hfill
	\begin{subfigure}[b]{0.7\textwidth}
		\centering
		\includegraphics[width=1\linewidth]{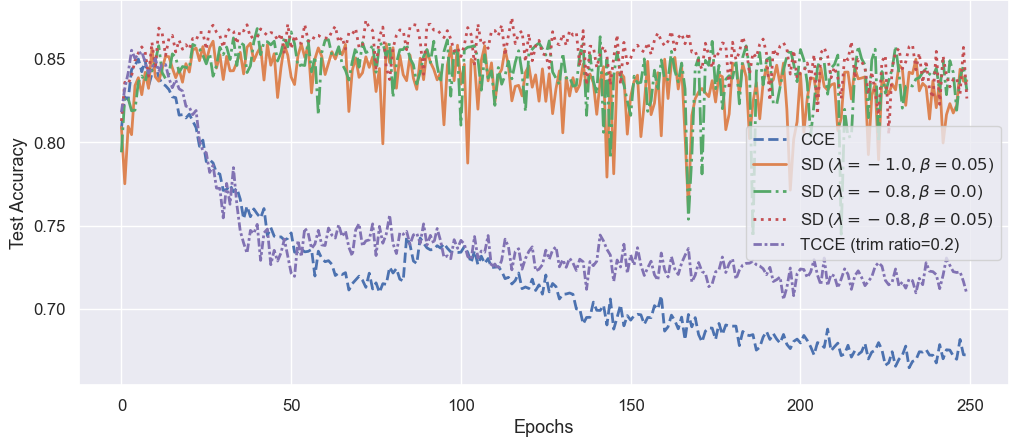}
		\caption{Contaminated training data with 40\% uniform label noise}
		\label{fig:mnist-history_plot_40}
	\end{subfigure}
	\hfill
	\begin{subfigure}[b]{0.7\textwidth}
		\centering
		\includegraphics[width=1\linewidth]{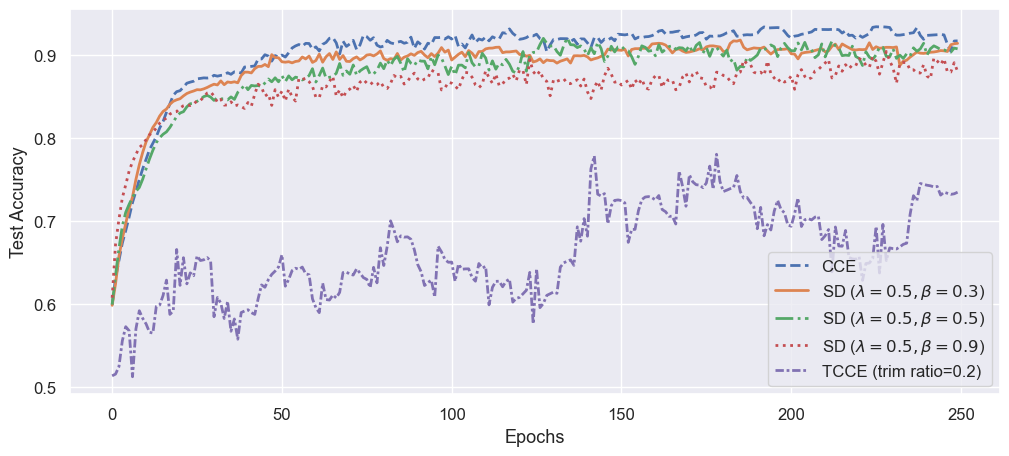}
		\caption{Contaminated training data under FGSM adversarial attack}
		\label{fig:mnist-history_plot_fgsm}
	\end{subfigure}
	\caption{Test accuracies obtained by different NN learning methods trained with varying numbers of epochs for the MNIST dataset}
	\label{fig:history_mnist}
\end{figure}

\begin{figure}[H]
	\centering
	\begin{subfigure}[b]{0.7\textwidth}
		\centering
		\includegraphics[width=1\linewidth]{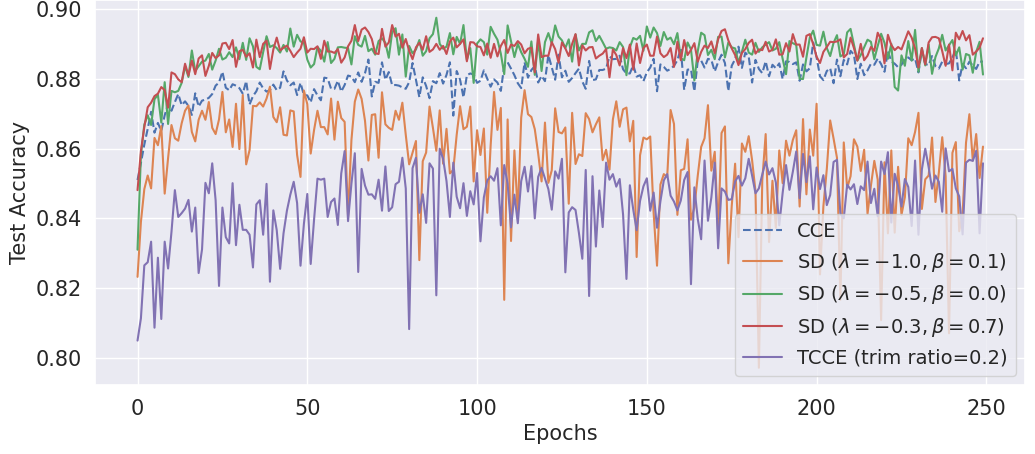}
		\caption{Clean training data}
		\label{fig:history_plot_0}
	\end{subfigure}
	\hfill
	\begin{subfigure}[b]{0.7\textwidth}
		\centering
		\includegraphics[width=1\linewidth]{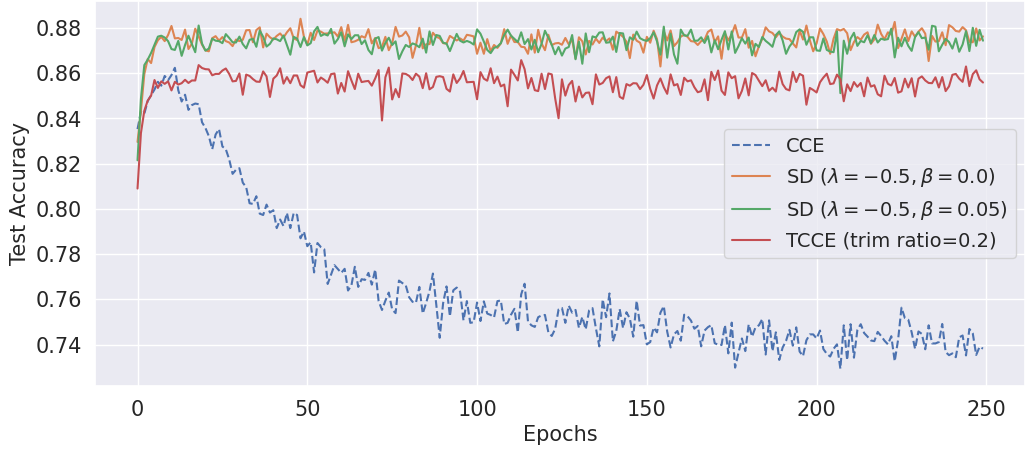}
		\caption{Contaminated training data with 20\% uniform label noise}
		\label{fig:history_plot_20}
	\end{subfigure}
	\hfill
	\begin{subfigure}[b]{0.7\textwidth}
		\centering
		\includegraphics[width=1\linewidth]{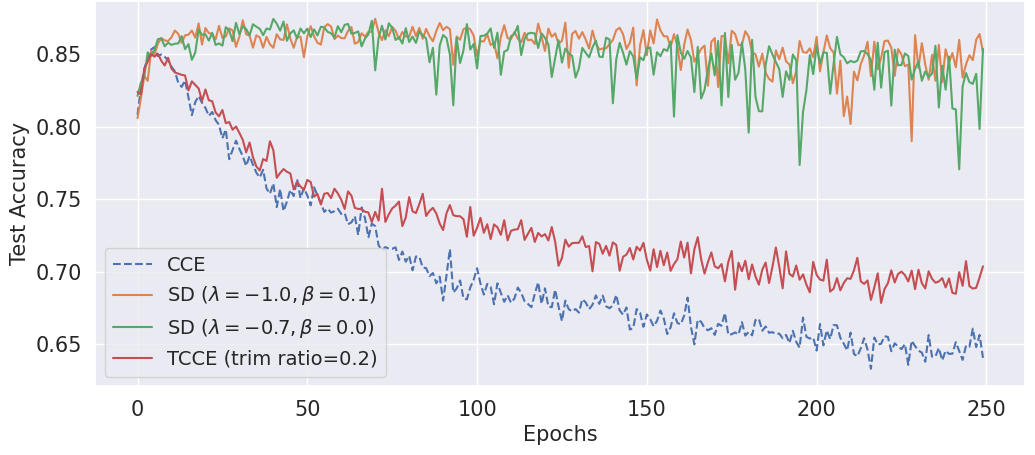}
		\caption{Contaminated training data with 40\% uniform label noise}
		\label{fig:history_plot_40}
	\end{subfigure}
	\caption{Test accuracies obtained by different NN learning methods trained with varying numbers of epochs for the Fashion-MNIST dataset}
	\label{fig:history_fmnist}
\end{figure}

Here we provide a brief look at the convergence behavior of the proposed rSDNet with respect to the numbers of training epochs,
as this directly determines the computational cost of applying robust rSDNet in practice. 
For this purpose, we refitted the same NN models described in Section \ref{SEC:numericals} on the MNIST and Fashion-MNIST datasets,
but now using their default training-test split as provided in \texttt{TensorFlow}. 
We then evaluate test accuracy for models trained over a range of epochs from 1 to 250. 
For both datasets, experiments were repeated for clean training data and contaminated training data with 20\% and 40\% uniform label noise.
Additionally, for the MNIST dataset, we also considered adversarially perturbed training data generated using FGSM attacks.
The resulting test accuracies are reported in Figures \ref{fig:history_mnist}--\ref{fig:history_fmnist} 
for rSDNet with a few representative values of $(\beta, \lambda)\in\mathcal{T}$,
alongside benchmark results obtained using the standard CCE loss and its trimmed variant TCCE(0.2).

The results show that, under clean data, rSDNet exhibits a convergence rate comparable to both CCE and TCCE(0.2), 
with all methods converging within a small number of epochs. 
However, in the presence of uniform label noise, CCE and TCCE display slower and less stable convergence, 
whereas rSDNet maintains a fast and stable convergence behavior similar to that observed in the clean-data scenario.
Under adversarial corruption as well, rSDNet achieves convergence speeds comparable to CCE, 
while TCCE-based methods converge significantly more slowly.
These findings demonstrate that rSDNet can be effectively applied to practical datasets,
whether clean or contaminated by structured noise, 
without incurring additional computational cost to achieve desired level of robustness.

\newpage
\section{Tables containing empirical results}\label{APP:tabs}
\begin{table}[H]
	\centering
	\caption{Average $k$-fold CV accuracies of the NN classifiers trained on the three selected datasets
    using the proposed rSDNet with different $(\beta, \lambda)$ and the benchmark losses/classifiers under clean data
[$k=7$ for MNIST and Fashion-MNIST, and $k=6$ for CIFAR-10]}
	\begin{tabular}{l|ccc}
		\hline
		Dataset $\rightarrow$ & MNIST & Fashion-MNIST & CIFAR-10 \\
		\hline
		\multicolumn{4}{l}{\underline{Existing losses for neural classification}} \\
		CCE & 0.9816 & 0.8891 & 0.6782 \\
		MAE & 0.9770 & 0.7940 & 0.1004 \\
		TCCE(0.1) & 0.9354 & 0.8815 & 0.6733 \\
		TCCE(0.2) & 0.8665 & 0.3094 & 0.6687 \\
		TCCE(0.3) & 0.7787 & 0.2939 & 0.6275 \\
		rKLD & 0.9643 & 0.8803 & 0.6240 \\
		SCE$(\alpha=0.5,\beta=1)$ & 0.9809 & 0.8911 & 0.6785 \\
		GCE$(q=0.5)$ & 0.9788 & 0.8932 & 0.6629 \\
		GCE$(q=0.7)$ & 0.9796 & 0.8764 & 0.5590 \\
		FCL$(\mu = 0)$ & 0.9810 & 0.8908 & 0.1000 \\
		FCL$(\mu = 0.25)$ & 0.9793 & 0.8898 & 0.6705 \\
		FCL$(\mu = 0.5)$ & 0.1014 & 0.1000 & 0.6657 \\
		FCL$(\mu = 0.75)$ & 0.1029 & 0.1007 & 0.5708 \\ \hline
		\\
		\multicolumn{4}{l}{\underline{Proposed rSDNet, with different $(\beta,\lambda)$}} \\
		$(0.05,-1)$ & 0.9772 & 0.8472 & 0.6457 \\
		$(0.1,-1)$ & 0.9782 & 0.8518 & 0.6445 \\
		$(0.3,-1)$ & 0.9787 & 0.8832 & 0.6617 \\
		$(0.5,-1)$ & 0.9780 & 0.8937 & 0.6597 \\
		$(0.7,-1)$ & 0.9787 & 0.8939 & 0.6585 \\
		\hline
		$(0,-0.8)$ & 0.9789 & 0.8608 & 0.1938 \\
		$(0.05,-0.8)$ & 0.9782 & 0.8652 & 0.6615 \\
		$(0.1,-0.8)$ & 0.9791 & 0.8755 & 0.6493 \\
		$(0.3,-0.8)$ & 0.9786 & 0.8871 & 0.6571 \\
		$(0.5,-0.8)$ & 0.9797 & 0.8920 & 0.6638 \\
		$(0.7,-0.8)$ & 0.9789 & 0.8929 & 0.6487 \\
		\hline
		$(0,-0.7)$ & 0.9789 & 0.8724 & 0.5653 \\
		$(0.05,-0.7)$ & 0.9784 & 0.8779 & 0.6625 \\
		$(0.1,-0.7)$ & 0.9787 & 0.8804 & 0.6623 \\
		$(0.3,-0.7)$ & 0.9795 & 0.8879 & 0.6464 \\
		$(0.5,-0.7)$ & 0.9788 & 0.8925 & 0.6561 \\
		$(0.7,-0.7)$ & 0.9798 & 0.8923 & 0.6563 \\
		\hline
		$(0,-0.5)$ & 0.9796 & 0.8934 & 0.6705 \\
		$(0.05,-0.5)$ & 0.9806 & 0.8941 & 0.6689 \\
		$(0.1,-0.5)$ & 0.9801 & 0.8948 & 0.6664 \\
		$(0.3,-0.5)$ & 0.9801 & 0.8945 & 0.6531 \\
		$(0.5,-0.5)$ & 0.9801 & 0.8942 & 0.6559 \\
		$(0.7,-0.5)$ & 0.9782 & 0.8941 & 0.6569 \\
		\hline
		$(0.1,0)$ & 0.9807 & 0.8911 & 0.6729 \\
		$(0.3,0)$ & 0.9809 & 0.8916 & 0.6670 \\
		$(0.5,0)$ & 0.9807 & 0.8937 & 0.6639 \\
		$(0.7,0)$ & 0.9798 & 0.8936 & 0.6604 \\
		$(1,0)$ & 0.9790 & 0.8933 & 0.6536 \\
		\hline
		$(0.5,0.5)$ & 0.9807 & 0.8901 & 0.6626 \\
		$(0.7,0.5)$ & 0.9793 & 0.8933 & 0.6636 \\
		\hline
	\end{tabular}
	\label{tab:no-cont-CV}
\end{table}

\begin{table}[H]
	\centering
	\caption{Average 7-fold CV accuracies of the NN classifiers trained on the MNIST data using the proposed rSDNet with different $(\beta, \lambda)$ 
		and the benchmark losses/classifiers under uniform label noise with contamination proportion $\eta$ [Highest accuracy in each case is highlighted with bold font]}
	\begin{tabular}{l|rrrrr}
		\hline
%		& \multicolumn{5}{c}{$\eta$}\\
	$\eta\rightarrow$ & 0.1 & 0.2 & 0.3 & 0.4 & 0.5 \\ \hline
        \multicolumn{6}{l}{\underline{Existing losses for neural classification}} \\
		CCE & 0.8691 & 0.7528 & 0.6461 & 0.5789 & 0.5201 \\
		MAE & 0.9750 & 0.9720 & 0.9691 & \textbf{0.9662} & \textbf{0.9588} \\
		TCCE(0.1) & 0.9585 & 0.8917 & 0.7654 & 0.6535 & 0.5555 \\
		TCCE(0.2) & 0.9500 & 0.9456 & 0.8678 & 0.7448 & 0.6196 \\
		TCCE(0.3) & 0.8764 & 0.9476 & 0.9282 & 0.8408 & 0.7057 \\
		rKLD & 0.9031 & 0.8541 & 0.8123 & 0.7547 & 0.6899 \\
		SCE$(\alpha=0.5,\beta=1)$ & 0.8856 & 0.7901 & 0.7099 & 0.6373 & 0.5655 \\
		GCE$(q=0.5)$ & 0.9766 & 0.9715 & 0.9424 & 0.8424 & 0.7300 \\
		GCE$(q=0.7)$ & 0.9758 & 0.9731 & 0.9694 & 0.9629 & 0.9507 \\
		FCL$(\mu=0)$ & 0.8762 & 0.7593 & 0.6608 & 0.6004 & 0.5207 \\
		FCL$(\mu=0.25)$ & 0.8796 & 0.7718 & 0.6790 & 0.6052 & 0.5393 \\
		FCL$(\mu=0.50)$ & 0.1437 & 0.8215 & 0.7258 & 0.6544 & 0.5846 \\
		FCL$(\mu=0.75)$ & 0.1095 & 0.1063 & 0.1016 & 0.1009 & 0.0983 \\ \hline
		\\
		\multicolumn{6}{l}{\underline{Proposed rSDNet, with different $(\beta,\lambda)$}} \\
		$(0.05,-1)$ & 0.9742 & 0.9731 & \textbf{0.9702} & 0.9613 & \textbf{0.9544} \\
		$(0.1,-1)$ & 0.9753 & 0.9726 & 0.9688 & 0.9624 & 0.9407 \\
		$(0.3,-1)$ & 0.9752 & 0.9683 & 0.8656 & 0.7260 & 0.6165 \\
		$(0.5,-1)$ & 0.9746 & 0.8866 & 0.7706 & 0.6682 & 0.5922 \\
		$(0.7,-1)$ & 0.9613 & 0.8565 & 0.7404 & 0.6558 & 0.5906 \\
		$(1,-1)$ & 0.9495 & 0.8451 & 0.7469 & 0.6491 & 0.5692 \\ \hline
		$(0   ,-0.8)$ & 0.9757 & \textbf{0.9743} & 0.9681 & \textbf{0.9646} & 0.9542 \\
		$(0.05,-0.8)$ & 0.9760 & 0.9728 & 0.9682 & 0.9618 & 0.9451 \\
		$(0.1 ,-0.8)$ & 0.9763 & 0.9734 & 0.9683 & 0.9578 & 0.8782 \\
		$(0.3 ,-0.8)$ & 0.9753 & 0.9677 & 0.8568 & 0.7372 & 0.6200 \\
		$(0.5 ,-0.8)$ & 0.9744 & 0.8917 & 0.7789 & 0.6803 & 0.5910 \\
		$(0.7 ,-0.8)$ & 0.9659 & 0.8598 & 0.7555 & 0.6593 & 0.5842 \\ \hline
		$(0   ,-0.7)$ & 0.9772 & 0.9732 & 0.9698 & 0.9619 & 0.9503 \\
		$(0.05,-0.7)$ & 0.9758 & 0.9732 & 0.9668 & 0.9611 & 0.9228 \\
		$(0.1 ,-0.7)$ & \textbf{0.9774} & 0.9737 & 0.9671 & 0.9465 & 0.8092 \\
		$(0.3 ,-0.7)$ & 0.9764 & 0.9608 & 0.8556 & 0.7360 & 0.6169 \\
		$(0.5 ,-0.7)$ & 0.9735 & 0.8932 & 0.7716 & 0.6763 & 0.5909 \\
		$(0.7 ,-0.7)$ & 0.9667 & 0.8605 & 0.7560 & 0.6703 & 0.5810 \\ \hline
		$(0  ,-0.5)$ & 0.9766 & 0.9704 & 0.9433 & 0.8466 & 0.7340 \\
		$(0.1,-0.5)$ & 0.9755 & 0.9684 & 0.9046 & 0.7893 & 0.6667 \\
		$(0.3,-0.5)$ & 0.9754 & 0.9302 & 0.8153 & 0.7139 & 0.5945 \\
		$(0.5,-0.5)$ & 0.9730 & 0.8927 & 0.7777 & 0.6847 & 0.5825 \\
		$(0.7,-0.5)$ & 0.9676 & 0.8695 & 0.7577 & 0.6599 & 0.5772 \\ \hline
		$(0.1,0)$ & 0.8689 & 0.7552 & 0.6565 & 0.5876 & 0.5228 \\
		$(0.3,0)$ & 0.9021 & 0.7906 & 0.6948 & 0.6127 & 0.5333 \\
		$(0.5,0)$ & 0.9439 & 0.8410 & 0.7299 & 0.6431 & 0.5598 \\
		$(0.7,0)$ & 0.9599 & 0.8642 & 0.7589 & 0.6565 & 0.5725 \\ \hline
		$(0.5,0.5)$ & 0.8837 & 0.7670 & 0.6745 & 0.5927 & 0.5320 \\
		$(0.7,0.5)$ & 0.9397 & 0.8440 & 0.7406 & 0.6448 & 0.5573 \\ \hline
	\end{tabular}
	\label{tab:mnist-cv}
\end{table}
% We further employed a trimmed version of the DPD-loss (TDPD) for this classification problem under uniform label noise. The TDPD loss is simply obtained by substituting $\lambda=0$ in \eqref{SD-CCE} and trimming a certain proportion of the largest individual losses within the summation. The corresponding average CV test accuracies are reported in Table \ref{tab:mnist-cv-tdpd} in the Appendix \ref{add-results}. The results show that the TDPD also provides a good, robust performance under low or moderate levels of contamination with appropriate hyperparameters.

\begin{table}[H]
	\centering
	\caption{Average 7-fold CV accuracies of the NN classifiers trained on the Fashion-MNIST data using the proposed rSDNet with different $(\beta, \lambda)$ 
		and the benchmark losses/classifiers under uniform label noise with contamination level $\eta$  [Highest accuracy in each case is highlighted with bold font]}
	\begin{tabular}{l|rrrrr}
		\hline
%		& \multicolumn{5}{c}{$\eta$}\\
	$\eta\rightarrow$ & 0.1 & 0.2 & 0.3 & 0.4 & 0.5 \\ \hline
        \multicolumn{6}{l}{\underline{Existing losses for neural classification}} \\
		CCE & 0.8015 & 0.7401 & 0.6955 & 0.6549 & 0.6205 \\
		MAE & 0.7910 & 0.7907 & 0.7224 & 0.7491 & 0.7179 \\
		TCCE(0.1) & 0.8771 & 0.8070 & 0.7387 & 0.6669 & 0.6098 \\ 
		TCCE(0.2) & 0.8769 & 0.8630 & 0.7923 & 0.7137 & 0.6327 \\ 
		TCCE(0.3) & 0.6118 & 0.8715 & 0.8509 & 0.7794 & 0.6848 \\
		rKLD & 0.8459 & 0.8099 & 0.7711 & 0.7276 & 0.6871 \\
		SCE$(\alpha=0.5,\beta=1)$ & 0.8241 & 0.7781 & 0.7189 & 0.6772 & 0.6235 \\
		GCE$(q=0.5)$ & 0.8868 & 0.8816 & 0.8557 & 0.7816 & 0.6756 \\
		GCE$(q=0.7)$ & 0.8688 & 0.8658 & 0.8531 & 0.8604 & 0.8339 \\
		FCL$(\mu=0)$ & 0.8125 & 0.7513 & 0.7028 & 0.6558 & 0.6177 \\
		FCL$(\mu=0.25)$ & 0.8189 & 0.7514 & 0.6990 & 0.6451 & 0.6084 \\
		FCL$(\mu=0.50)$ & 0.8051 & 0.6845 & 0.7202 & 0.6644 & 0.6123 \\
		FCL$(\mu=0.75)$ & 0.0978 & 0.0990 & 0.0989 & 0.1000 & 0.0994 \\ \hline
		\\
		\multicolumn{6}{l}{\underline{Proposed rSDNet, with different $(\beta,\lambda)$}} \\
		$(0.05,-1)$ & 0.8581 & 0.8381 & 0.8478 & 0.8364 & 0.8259 \\ 
		$(0.1,-1)$ & 0.8565 & 0.8480 & 0.8533 & \textbf{0.8629} & \textbf{0.8405} \\ 
		$(0.3,-1)$ & 0.8865 & 0.8651 & 0.7783 & 0.6881 & 0.6301 \\ 
		$(0.5,-1)$ & 0.8757 & 0.7928 & 0.7316 & 0.6812 & 0.6535 \\ 
		$(0.7,-1)$ & 0.8678 & 0.7890 & 0.7309 & 0.6777 & 0.6372 \\ \hline
		$(0,-0.8)$ & 0.8615 & 0.8494 & 0.8453 & 0.8365 & 0.8316 \\
		$(0.05,-0.8)$ & 0.8612 & 0.8617 & 0.8427 & 0.8473 & 0.8436 \\
		$(0.1,-0.8)$ & 0.8684 & 0.8649 & 0.8637 & 0.8623 & 0.7924 \\
		$(0.3,-0.8)$ & 0.8884 & 0.8648 & 0.7719 & 0.6941 & 0.6394 \\
		$(0.5,-0.8)$ & 0.8798 & 0.8044 & 0.7327 & 0.6889 & 0.6381 \\
		$(0.7,-0.8)$ & 0.8687 & 0.7920 & 0.7237 & 0.6822 & 0.6509 \\ \hline
		$(0,-0.7)$ & 0.8694 & 0.8640 & 0.8353 & 0.8583 & 0.8340 \\ 
		$(0.05,-0.7)$ & 0.8706 & 0.8717 & 0.8682 & 0.8509 & 0.8304 \\ 
		$(0.1,-0.7)$ & 0.8785 & 0.8758 & \textbf{0.8745} & 0.8468 & 0.7392 \\ 
		$(0.3,-0.7)$ & 0.8883 & 0.8631 & 0.7643 & 0.6903 & 0.6466 \\ 
		$(0.5,-0.7)$ & 0.8772 & 0.8061 & 0.7335 & 0.6821 & 0.6362 \\ 
		$(0.7,-0.7)$ & 0.8721 & 0.7877 & 0.7269 & 0.6837 & 0.6385 \\ \hline
		$(0,-0.5)$ & 0.8888 & \textbf{0.8832} & 0.8508 & 0.7728 & 0.6851 \\ 
		$(0.05,-0.5)$ & \textbf{0.8892} & 0.8788 & 0.8359 & 0.7593 & 0.6520 \\ 
		$(0.1,-0.5)$ & 0.8873 & 0.8749 & 0.8181 & 0.7268 & 0.6205 \\ 
		$(0.3,-0.5)$ & 0.8846 & 0.8440 & 0.7469 & 0.6776 & 0.6205 \\ 
		$(0.5,-0.5)$ & 0.8825 & 0.8085 & 0.7286 & 0.6785 & 0.6438 \\ 
		$(0.7,-0.5)$ & 0.8731 & 0.7931 & 0.7346 & 0.6719 & 0.6465 \\ \hline
		$(0.1,0)$ & 0.8037 & 0.7435 & 0.6922 & 0.6534 & 0.6093 \\ 
		$(0.3,0)$ & 0.8283 & 0.7541 & 0.6975 & 0.6587 & 0.6162 \\ 
		$(0.5,0)$ & 0.8542 & 0.7785 & 0.7083 & 0.6633 & 0.6377 \\ 
		$(0.7,0)$ & 0.8691 & 0.7848 & 0.7247 & 0.6782 & 0.6197 \\ 
		$(1,0)$ & 0.8649 & 0.7979 & 0.7240 & 0.6686 & 0.6272 \\ \hline
		$(0.5,0.5)$ & 0.8138 & 0.7425 & 0.6976 & 0.6525 & 0.6042 \\ 
		$(0.7,0.5)$ & 0.8535 & 0.7785 & 0.7167 & 0.6630 & 0.6170 \\ \hline
	\end{tabular}
	\label{tab:fmnist-cv}
\end{table}

\begin{table}[H]
	\centering
	\caption{Average 6-fold CV accuracies of the NN classifiers trained on the CIFAR-10 data using the proposed rSDNet with different $(\beta, \lambda)$ 
		and the benchmark losses/classifiers under uniform label noise with contamination proportion $\eta$  [Highest accuracy in each case is highlighted with bold font]}
	\begin{tabular}{l|rrrrr}
		\hline
%		& \multicolumn{5}{c}{$\eta$}\\
	$\eta\rightarrow$ & 0.1 & 0.2 & 0.3 & 0.4 & 0.5 \\ \hline
        \multicolumn{6}{l}{\underline{Existing losses for neural classification}} \\
		CCE & 0.5884 & 0.5117 & 0.4330 & 0.3696 & 0.3004 \\
		MAE & 0.0997 & 0.0988 & 0.1006 & 0.0987 & 0.0996 \\
		TCCE(0.1) & 0.6440 & 0.5816 & 0.5075 & 0.4243 & 0.3425 \\
		TCCE(0.2) & \textbf{0.6508} & 0.6137 & 0.5476 & 0.4665 & 0.3822 \\
		TCCE(0.3) & 0.6341 & 0.6120 & 0.5576 & 0.4896 & 0.3938 \\
		rKLD & 0.5445 & 0.4753 & 0.4024 & 0.3401 & 0.2695 \\		
		SCE$(\alpha=0.5,\beta=1)$ & 0.6013 & 0.5314 & 0.4517 & 0.3892 & 0.3141 \\
		GCE$(q=0.5)$ & 0.6470 & 0.5985 & 0.5429 & 0.4680 & 0.3676 \\
		GCE$(q=0.7)$ & 0.5532 & 0.5299 & 0.5021 & 0.4678 & 0.4070 \\
		FCL$(\mu = 0)$ & 0.1000 & 0.1000 & 0.1000 & 0.1000 & 0.1000 \\
		FCL$(\mu = 0.25)$ & 0.6087 & 0.5278 & 0.4468 & 0.3744 & 0.3075 \\
		FCL$(\mu = 0.5)$ & 0.6372 & 0.6014 & 0.5412 & 0.4539 & 0.3561 \\
		FCL$(\mu = 0.75)$ & 0.4519 & 0.1861 & 0.5030 & 0.3904 & 0.4054 \\ \hline
		\\
		\multicolumn{6}{l}{\underline{Proposed rSDNet, with different $(\beta,\lambda)$}} \\
		$(0.05,-1)$  & 0.3688 & 0.5142 & 0.5066 & \textbf{0.5385} & \textbf{0.4717} \\
		$(0.1,-1)$ & 0.6221 & 0.6063 & \textbf{0.5712} & 0.5103 & 0.4197 \\
		$(0.3,-1)$ & 0.6214 & 0.5707 & 0.4995 & 0.4024 & 0.3289 \\
		$(0.5,-1)$ & 0.6126 & 0.5499 & 0.4674 & 0.3838 & 0.3128 \\
		$(0.7,-1)$ & 0.6210 & 0.5468 & 0.4567 & 0.3834 & 0.3142 \\ \hline
		$(0,-0.8)$ & 0.3671 & 0.2676 & 0.4258 & 0.3272 & 0.3578 \\
		$(0.05,-0.8)$  & 0.6323 & \textbf{0.6153} & 0.5704 & 0.5269 & 0.4476 \\
		$(0.1,-0.8)$ & 0.6377 & 0.6083 & 0.5642 & 0.4941 & 0.3947 \\
		$(0.3,-0.8)$ & 0.6253 & 0.5726 & 0.5016 & 0.4058 & 0.3259 \\
		$(0.5,-0.8)$ & 0.6191 & 0.5530 & 0.4678 & 0.3920 & 0.3117 \\
		$(0.7,-0.8)$ & 0.6138 & 0.5444 & 0.4610 & 0.3842 & 0.3074 \\ \hline
		$(0,-0.7)$ & 0.4519 & 0.5311 & 0.4998 & 0.4653 & 0.4631 \\
		$(0.05,-0.7)$  & 0.6397 & 0.6009 & 0.5639 & 0.5119 & 0.4201 \\
		$(0.1,-0.7)$ & 0.6368 & 0.6061 & 0.5471 & 0.4721 & 0.3804 \\
		$(0.3,-0.7)$ & 0.6203 & 0.5730 & 0.4863 & 0.4140 & 0.3168 \\
		$(0.5,-0.7)$ & 0.6240 & 0.5465 & 0.4719 & 0.3920 & 0.3066 \\
		$(0.7,-0.7)$ & 0.6125 & 0.5453 & 0.4691 & 0.3857 & 0.3138 \\ \hline
		$(0,-0.5)$ & 0.6408 & 0.5977 & 0.5342 & 0.4632 & 0.3624 \\
		$(0.05,-0.5)$  & 0.6375 & 0.5963 & 0.5367 & 0.4403 & 0.3544 \\
		$(0.1,-0.5)$ & 0.6361 & 0.5920 & 0.5207 & 0.4284 & 0.3372 \\
		$(0.3,-0.5)$ & 0.6270 & 0.5644 & 0.4853 & 0.3971 & 0.3127 \\
		$(0.5,-0.5)$ & 0.6168 & 0.5513 & 0.4689 & 0.3934 & 0.3092 \\
		$(0.7,-0.5)$ & 0.6151 & 0.5477 & 0.4574 & 0.3890 & 0.3164 \\ \hline
		$(0.1,0)$ & 0.5929 & 0.5128 & 0.4378 & 0.3719 & 0.2974 \\
		$(0.3,0)$ & 0.5952 & 0.5113 & 0.4483 & 0.3701 & 0.2952 \\
		$(0.5,0)$ & 0.6079 & 0.5357 & 0.4489 & 0.3759 & 0.3047 \\
		$(0.7,0)$ & 0.6172 & 0.5390 & 0.4628 & 0.3798 & 0.3093 \\
		$(1,0)$ & 0.6163 & 0.5481 & 0.4594 & 0.3853 & 0.3124 \\ \hline
		$(0.5,0.5)$ & 0.5901 & 0.5126 & 0.4338 & 0.3703 & 0.2949 \\
		$(0.7,0.5)$ & 0.6099 & 0.5366 & 0.4527 & 0.3818 & 0.3042 \\ \hline
	\end{tabular}
	\label{tab:cifar10-cv}
\end{table}

\begin{table}[H]
	\centering
	\caption{Average test accuracies of the NN classifiers trained on adversarially perturbed MNIST images 
		using the proposed rSDNet with different $(\beta, \lambda)$ and the benchmark losses/classifiers 
		[Highest accuracy in each case is highlighted with bold font]}
	\begin{tabular}{l|rr|rr|rr|rr}
		\hline
	Attack type $\rightarrow$	& \multicolumn{2}{c|}{FGSM} & \multicolumn{2}{c|}{PGD} & \multicolumn{2}{c|}{CW} & \multicolumn{2}{c}{Deepfool} \\ \hline
		~ & \makecell{Clean\\test} & \makecell{Adv.\\test} & \makecell{Clean\\test} & \makecell{Adv.\\test} & \makecell{Clean\\test} & \makecell{Adv.\\test} & \makecell{Clean\\test} & \makecell{Adv.\\test} \\ \hline
        \multicolumn{9}{l}{\underline{Existing losses for neural classification}} \\
		CCE & 0.9073 & 0.9776 & 0.6516 & 0.9949 & \textbf{0.9815} & \textbf{0.9815} & 0.9731 & 0.9833 \\
		% \multicolumn{9}{l}{\underline{TCCE with trim ratio}} \\
        MAE & 0.9000 & 0.9744 & 0.6005 & 0.9910 & 0.9770 & 0.9770 & 0.9671 & 0.9807 \\
		TCCE(0.1) & 0.7947 & 0.9383 & 0.4653 & 0.9870 & 0.9549 & 0.9549 & 0.9466 & 0.9574 \\
		TCCE(0.2) & 0.7387 & 0.9117 & 0.4493 & 0.9434 & 0.9184 & 0.9184 & 0.9086 & 0.9147 \\
		TCCE(0.3) & 0.6482 & 0.8421 & 0.4142 & 0.8823 & 0.8808 & 0.8808 & 0.8848 & 0.8733 \\
        
        SCE$(\alpha=0.5,\beta=1)$ & 0.9094 & 0.9765 & \textbf{0.6732} & 0.9948 & 0.9809 & 0.9809 & 0.9720 & 0.9846 \\
        GCE($q=0.5$) & 0.9133 & 0.9730 & 0.5858 & 0.9913 & 0.9788 & 0.9788 & 0.9708 & 0.9841 \\
        GCE($q=0.7$) & 0.9090 & 0.9745 & 0.5992 & 0.9910 & 0.9796 & 0.9796 & 0.9666 & 0.9811 \\
        FCL$(\mu=0)$ & 0.9156 & 0.9773 & 0.6777 & 0.9945 & 0.9810 & 0.9810 & 0.9735 & 0.9844 \\
        FCL$(\mu=0.25)$ & 0.9074 & 0.9733 & 0.6288 & 0.9684 & 0.9793 & 0.9793 & 0.9709 & 0.9831 \\
        FCL$(\mu=0.5)$ & 0.0999 & 0.0999 & 0.1012 & 0.1012 & 0.1014 & 0.1014 & 0.1016 & 0.1016 \\
        FCL$(\mu=0.75)$ & 0.1003 & 0.1003 & 0.1052 & 0.1052 & 0.1029 & 0.1029 & 0.1025 & 0.1025 \\ \hline
        \\
		\multicolumn{9}{l}{\underline{Proposed rSDNet, with different $(\beta,\lambda)$}} \\
		% \multicolumn{9}{l}{\underline{SD with $\lambda=-1$ and $\beta$}} \\
		$(0.1,-1)$ & 0.9080 & 0.9726 & 0.5822 & 0.9957 & 0.9774 & 0.9774 & 0.9678 & 0.9821 \\
		$(0.3,-1)$ & 0.9083 & 0.9743 & 0.5848 & \textbf{0.9958} & 0.9793 & 0.9793 & 0.9664 & 0.9812 \\
		$(0.5,-1)$ & 0.9221 & 0.9749 & 0.6157 & 0.9955 & 0.9788 & 0.9788 & 0.9695 & 0.9830 \\
		$(0.7,-1)$ & 0.9194 & 0.9753 & 0.6174 & 0.9954 & 0.9781 & 0.9781 & 0.9674 & 0.9814 \\
		$(0.9,-1)$ & 0.9192 & 0.9752 & 0.5984 & 0.9954 & 0.9791 & 0.9791 & 0.9708 & 0.9835 \\  \hline
		% \multicolumn{9}{l}{\underline{SD with $\lambda=-0.7$ and $\beta$}} \\
		$(0.1,-0.7)$ & 0.9078 & 0.9750 & 0.5879 & 0.9954 & 0.9788 & 0.9788 & 0.9691 & 0.9823 \\
		$(0.3,-0.7)$ & 0.9112 & 0.9738 & 0.5951 & 0.9953 & 0.9784 & 0.9784 & 0.9661 & 0.9808 \\
		$(0.5,-0.7)$ & 0.9141 & 0.9753 & 0.5968 & 0.9955 & 0.9790 & 0.9790 & 0.9686 & 0.9826 \\
		$(0.7,-0.7)$ & 0.9184 & 0.9748 & 0.6068 & 0.9953 & 0.9792 & 0.9792 & 0.9675 & 0.9817 \\
		$(0.9,-0.7)$ & 0.9214 & 0.9747 & 0.6004 & 0.9957 & 0.9795 & 0.9795 & 0.9683 & 0.9819 \\  \hline
		% \multicolumn{9}{l}{\underline{SD with $\lambda=-0.5$ and $\beta$}} \\
		$(0.1,-0.5)$ & 0.9105 & 0.9749 & 0.5940 & 0.9953 & 0.9787 & 0.9787 & 0.9703 & 0.9838 \\
		$(0.3,-0.5)$ & 0.9122 & 0.9756 & 0.6017 & 0.9956 & 0.9796 & 0.9796 & 0.9685 & 0.9822 \\
		$(0.5,-0.5)$ & 0.9138 & 0.9759 & 0.6030 & 0.9952 & 0.9789 & 0.9789 & 0.9688 & 0.9825 \\
		$(0.7,-0.5)$ & 0.9205 & 0.9751 & 0.6081 & 0.9954 & 0.9797 & 0.9797 & 0.9683 & 0.9818 \\
		$(0.9,-0.5)$ & 0.9186 & 0.9753 & 0.6018 & 0.9955 & 0.9785 & 0.9785 & 0.9683 & 0.9823 \\  \hline
		% \multicolumn{9}{l}{\underline{SD with $\lambda=-0.3$ and $\beta$}} \\
		$(0.1,-0.3)$ & 0.9099 & 0.9775 & 0.5743 & 0.9956 & 0.9797 & 0.9797 & 0.9712 & 0.9844 \\
		$(0.3,-0.3)$ & 0.9110 & 0.9763 & 0.5820 & 0.9954 & 0.9787 & 0.9787 & 0.9720 & 0.9847 \\
		$(0.5,-0.3)$ & 0.9098 & 0.9757 & 0.5796 & 0.9953 & 0.9794 & 0.9794 & 0.9706 & 0.9834 \\
		$(0.7,-0.3)$ & 0.9209 & 0.9749 & 0.5900 & 0.9955 & 0.9792 & 0.9792 & 0.9688 & 0.9826 \\
		$(0.9,-0.3)$ & 0.9143 & 0.9764 & 0.6153 & 0.9955 & 0.9789 & 0.9789 & 0.9696 & 0.9829 \\  \hline
		% \multicolumn{9}{l}{\underline{SD with $\lambda=0$ and $\beta$ (DPD)}} \\
		$(0.1,0)$ & 0.9125 & 0.9768 & 0.6265 & 0.9949 & 0.9807 & 0.9807 & 0.9730 & 0.9844 \\
		$(0.3,0)$ & 0.9124 & 0.9763 & 0.6439 & 0.9950 & 0.9808 & 0.9808 & 0.9726 & 0.9853 \\
		$(0.5,0)$ & 0.9188 & 0.9741 & 0.5982 & 0.9955 & 0.9793 & 0.9793 & 0.9713 & 0.9846 \\
		$(0.7,0)$ & 0.9108 & 0.9760 & 0.5969 & 0.9952 & 0.9787 & 0.9787 & 0.9707 & 0.9837 \\
		$(1  ,0)$ & 0.9160 & 0.9755 & 0.5865 & 0.9955 & 0.9788 & 0.9788 & 0.9680 & 0.9823 \\  \hline
		% \multicolumn{9}{l}{\underline{SD with $\lambda=0.5$ and $\beta$}} \\
		$(0.1,0.5)$ & 0.8725 & 0.9747 & 0.5922 & 0.9944 & 0.9702 & 0.9702 & 0.9643 & 0.9771 \\
		$(0.3,0.5)$ & 0.9076 & \textbf{0.9779} & 0.6222 & 0.9947 & 0.9809 & 0.9809 & 0.9727 & 0.9846 \\
		$(0.5,0.5)$ & 0.9116 & 0.9755 & 0.6273 & 0.9952 & 0.9809 & 0.9809 & \textbf{0.9732} & \textbf{0.9854} \\
		$(0.7,0.5)$ & 0.9123 & 0.9755 & 0.6176 & 0.9948 & 0.9805 & 0.9805 & 0.9708 & 0.9841 \\
		$(0.9,0.5)$ & \textbf{0.9235} & 0.9757 & 0.5955 & 0.9956 & 0.9785 & 0.9785 & 0.9690 & 0.9822 \\  \hline
	\end{tabular}
	\label{tab:adv-attack}
\end{table}

\bibliographystyle{apalike}
\bibliography{Reference.bib}
\end{document}